\documentclass[10pt,journal,letterpaper,twoside,twocolumn]{IEEEtran}
\usepackage{amsmath,graphicx}
\usepackage[utf8]{inputenc}
\usepackage{psfrag,epsfig,graphics}
\usepackage{amsmath,amsthm,amssymb,multirow}
\usepackage{mathbbol}
\usepackage{amssymb}             
\usepackage{mathrsfs}

\usepackage{mathabx} 

\DeclareSymbolFontAlphabet{\amsmathbb}{AMSb}%

\usepackage{booktabs}
\usepackage{graphicx}
\usepackage{caption}
\usepackage[position=b]{subcaption}

\usepackage[noadjust]{cite}
\usepackage{multirow}
\usepackage[lined,linesnumbered,ruled]{algorithm2e}

\usepackage{color}  
\def\cred{\textcolor{red}}

\def\cblue{\textcolor{blue}}

\usepackage[colorinlistoftodos]{todonotes}  

\newtheorem{definition}{Definition}

\newcommand{\cb}[1]{{\boldsymbol{#1}}}
\newcommand{\cp}[1]{\ifmmode {\mathcal{#1}}\else ${\mathcal{#1}}$\fi}

\newcommand{\bB}{\boldsymbol{B}}

\newcommand{\bI}{\boldsymbol{I}}

\newcommand{\bM}{\boldsymbol{M}}

\newcommand{\bX}{\boldsymbol{X}}

\newcommand{\bZ}{\boldsymbol{Z}}

\newcommand{\bm}{\boldsymbol{m}}

\newcommand{\be}{\boldsymbol{e}}

\newcommand{\br}{\boldsymbol{r}}

\newcommand{\bx}{\boldsymbol{x}}
\newcommand{\bz}{\boldsymbol{z}}

\newcommand{\tensor}[1]{\cb{\mathcal{#1}}}

\newcommand{\balpha}{\boldsymbol{\alpha}}

\newcommand{\bbPsi}{\mathbb{\Psi}}
\newcommand{\bbLambda}{\tensor{P}}

\title{Low-Rank Tensor Modeling for Hyperspectral Unmixing Accounting for Spectral Variability }
\author{Tales Imbiriba,~\IEEEmembership{Member,~IEEE},
Ricardo~Augusto~Borsoi,~\IEEEmembership{Student Member,~IEEE}, 
 Jos\'e~Carlos~Moreira~Bermudez,~\IEEEmembership{Senior~Member,~IEEE}
\thanks{This work has been supported by the National Council for Scientific and Technological Development (CNPq) under grants 304250/2017-1, 409044/2018-0, 141271/2017-5 and 204991/2018-8, and by the Brazilian Education Ministry (CAPES) under grant PNPD/1811213.}
\thanks{T. Imbiriba was with the Department of Electrical Engineering, Federal University of Santa Catarina (DEE--UFSC), Florian\'opolis, SC, Brazil, and is with the ECE department of the Northeastern University, Boston, MA, USA. e-mail: \mbox{talesim@ece.neu.edu}.
R.A. Borsoi is with the DEE--UFSC, Florian\'opolis, SC, Brazil, and with the Lagrange Laboratory, Universit\'e  C\^ote  d'Azur, Nice, France. e-mail: \mbox{raborsoi@gmail.com}.
J.C.M. Bermudez is with the DEE--UFSC, Florian\'opolis, SC, Brazil, and with the Graduate Program on Electronic Engineering and Computing, Catholic University of Pelotas (UCPel) Pelotas, Brazil. e-mail: \mbox{j.bermudez@ieee.org}.}
\thanks{Manuscript received Month day, year; revised Month day, year.}}

\begin{document}
%

\markboth{ACCEPTED for publication in the IEEE TRANSACTIONS ON GEOSCIENCE AND REMOTE SENSING --
~\today.}{Imbiriba, Borsoim, Bermudez} 
\maketitle
\begin{abstract}
Traditional hyperspectral unmixing methods neglect the underlying variability of spectral signatures often observed in typical hyperspectral images (HI), propagating these missmodeling errors throughout the whole unmixing process. Attempts to model material spectra as members of sets or as random variables tend to lead to severely ill-posed unmixing problems.
Although parametric models have been proposed to overcome this drawback by handling endmember variability through generalizations of the mixing model, the success of these techniques depend on employing appropriate regularization strategies. Moreover, the existing approaches fail to adequately explore the natural multidimensinal representation of HIs.
Recently, tensor-based strategies considered low-rank decompositions of hyperspectral images as an alternative to impose low-dimensional structures on the solutions of standard and multitemporal unmixing problems
. These strategies, however, present two main drawbacks: 1) they confine the solutions to low-rank tensors, which often cannot represent the complexity of real-world scenarios; and 2) they lack guarantees that endmembers and abundances will be correctly factorized in their respective tensors. In this work, we propose a more flexible approach, called ULTRA-V, that imposes low-rank structures through regularizations whose strictness is controlled by scalar parameters. Simulations attest the superior accuracy of the method when compared with state-of-the-art unmixing algorithms that account for spectral variability.
\end{abstract}
\section{Introduction}
\label{sec:intro}

Hyperspectral imaging has attracted formidable interest from the scientific community in the past two decades, and hyperspectral images (HI) have been explored in a vast, and increasing, number of applications in different fields~\cite{Bioucas-Dias-2013-ID307}. The limited spatial resolution of hyperspectral devices often mixes the spectral contributions of different pure materials, termed \emph{endmembers} (EM), in the scene~\cite{Keshava:2002p5667}. This phenomenon is more explicit in remote sense applications, due to the distance between airborne and spaceborne sensors and the target scene. The mixing process must be well understood for the vital information relating the pure materials and their distribution in the scene to be accurately unveiled. 
Hyperspectral unmixing (HU) aims at solving this problem by decomposing the hyperspectral image into a collection of endmembers and their fractional \emph{abundances}~\cite{Bioucas2012}. 

Different mixing models have been employed to explain the interaction between light and the endmembers~\cite{Bioucas-Dias-2013-ID307}-\cite{Dobigeon-2014-ID322}.
The simplest and most widely used model is the Linear Mixing Model (LMM)~\cite{Keshava:2002p5667}, which assumes that the observed reflectance vector (\emph{i.e.} a pixel) can be modeled as a convex combination of the spectral signatures of each endmember present in the scene. Convexity imposes positivity and sum-to-one constraints on the linear combination coefficients. Hence,  they represent the fractional abundances with which the endmembers contribute to the scene. Though the simplicity of the LMM leads to fast and reliable unmixing strategies in some situations, it turns out to be simplistic to explain the mixing process in many practical applications. Hence, several approaches have been proposed in the literature to account for nonlinear mixing effects~\cite{Dobigeon-2014-ID322,Imbiriba2016_tip, Imbiriba2017_bs_tip} and endmember variability~\cite{Zare-2014-ID324, drumetz2016variabilityReviewRecent,borsoi2018superResolution} often present in practical scenes.

A myriad of factors can induce endmember variability, including environmental, illumination, atmospheric and temporal changes~\cite{Zare-2014-ID324}. If not properly considered, such variability can result in significant estimation errors being propagated throughout the unmixing process~\cite{Thouvenin_IEEE_TSP_2016}. Most of the methods proposed so far to deal with spectral variability can be classified in three major groups: endmembers as sets, endmembers as statistical distributions and, more recently, methods that incorporate the variability in the mixing model, often using physically motivated concepts~\cite{drumetz2016variabilityReviewRecent}. The method proposed in this work belongs to the third group. Parametric models have received considerable interest since, unlike the other two approaches, they require neither spectral libraries to be know a priori nor strong hypothesis about the endmembers statistical distribution.

Recently, \cite{Thouvenin_IEEE_TSP_2016}, \cite{drumetz2016blind} and \cite{Imbiriba_glmm_2018} introduced variations of the LMM to cope with the spectral variability. Unmixing using these models lead to ill-posed problems that were solved by using a combination of different regularizations terms and variable splitting optimization strategies. The model Perturbed LMM (PLMM) in~\cite{Thouvenin_IEEE_TSP_2016} augmented the endmember matrix with an additive perturbation matrix that needs to be estimated jointly with the abundances. Although the additive perturbation can model arbitrary endmember variations, it is not physically motivated, and the excessive amount of degrees of freedom makes the problem even harder to solve. The Extended LMM (ELMM) proposed in~\cite{drumetz2016blind} introduces one new multiplicative term for each endmember, and can efficiently model changes in the observed reflectance due to illumination effects~\cite{drumetz2016blind}. This model has a clear physical motivation, but its modeling capability is limited. The Generalized Linear Mixing Model (GLMM) proposed in~\cite{Imbiriba_glmm_2018} generalizes the ELMM to account for variability in all regions of the measured spectrum. The GLMM is physically motivated and capable of modeling arbitrary variability, resulting in improved accuracy at the expense of a small increase in the computational complexity, when compared to the ELMM. Other works attempt to capture these complex spectral variations indirectly by means of additive residual terms~\cite{Halimi_IEEE_Trans_CI_2017,hong2019augmentedLMMvariability}. Although avoiding the interactions between the abundance fractions and the endmember signatures, these strategies usually do not estimate the EM spectra for each image pixel.

The above mentioned methods resort to different strategies to regularize the ill-posed optimization problem leading to the estimation of abundances and endmembers. The regularization is achieved by introducing into the unmixing problem additional information based on common knowledge about the low-dimensionality of structures embedded in hyperspectral images.

Possible ways to recover lower-dimensional structures from noisy and corrupted data include the imposition of low-rank matrix constraints on the estimation process~\cite{tao2011recovering}, or the low-rank decomposition of the observed data~\cite{bousse2017tensor,sidiropoulos2017tensor}. 
The facts that HIs are naturally represented and treated as tensors, and that low-rank decompositions of higher-order ($>$2) tensors tend to capture homogeneities within the tensor structure make such strategies even more attractive for HU.
Low-rank tensor models have been successfully employed in various tasks involving HIs, such as recovery of missing pixels~\cite{ng2017tensorHSIrecoveryMissing}, anomaly detection~\cite{zhang2016tensorAnomalyDetectionHSI}, classification~\cite{guo2016supportTensorMachines}, compression~\cite{yang2015tensorHSIcompression}, dimensionality reduction~\cite{zhang2013tensorDimensionalityReductionHSI} and analysis of multi-angle images~\cite{veganzones2016tensorCPdecHSImultiangle}.
More recently, \cite{qian2017tensorNMFunmixing} and \cite{veganzones2016tensorCPdecHSImultiangle} considered low-rank tensor decompositions applied to standard and multitemporal HU, respectively. 

In~\cite{qian2017tensorNMFunmixing} the HI is treated as three-dimensional tensor, and spatial regularity is enforced through a nonnegative tensor factorization (NTF) strategy that imposes a low-rank tensor structure.
In~\cite{veganzones2016tensorCPdecHSImultiangle}, nonnegative canonical polyadic decomposition were used to unmix multitemporal HIs represented as three-dimensional tensors built by stacking multiple temporal matricized HIs.
Though a low-rank tensor representation may naturally describe the regularity of HIs and abundance maps, the forceful introduction of stringent rank constraints may prevent an adequate representation of fast varying structures that are important for accurate unmixing.
Another limitation of the approaches proposed in~\cite{qian2017tensorNMFunmixing} and~\cite{veganzones2016tensorCPdecHSImultiangle} is the lack of guarantee that endmembers and abundances will be correctly factorized into their respective tensors. In~\cite{imbiriba2018_ULTRA}, we proposed a new low-rank HU method called \textit{Unmixing with Low-rank Tensor Regularization Algorithm} (ULTRA), which accounts for highly correlated endmembers. The HU problem was formulated using tensors and a low-rank abundance tensor regularization term was introduced. Differently, from the strict tensor decomposition considered in~\cite{qian2017tensorNMFunmixing,veganzones2016tensorCPdecHSImultiangle}, ULTRA allowed important flexibility to the rank of the estimated abundance tensor to adequately represent fine scale structure and details that lie beyond a low-rank structure, but without compromising the regularity of the solution.

In this work we extend the strategy proposed in~\cite{imbiriba2018_ULTRA} to account for the important effect of endmember variability as well as a novel method to estimate the sufficient rank of a tensor for accurately solving the HI unmixing problem. The main novel contributions of this paper are:
\begin{itemize}
 \item[a)] We extend the strategy proposed in~\cite{imbiriba2018_ULTRA} by imposing a new low-rank regularization on the four-dimensional endmember tensor, which contains one endmember matrix for each pixel, to account for endmember variability. The new cost function results in an iterative algorithm, named \textit{Unmixing with Low-rank Tensor Regularization Algorithm accounting for endmember Variability} (ULTRA-V).  At each iteration, ULTRA-V updates the estimations of the abundance and endmember tensors as well as their low-rank approximations.
 \item[b)] We propose a novel non-trivial strategy to determine the smallest rank representation that contains most of the variation of multilinear singular values~\cite{de2000multilinear}. 
\end{itemize}
Simulation results using synthetic and real data illustrate the performance improvement obtained using ULTRA-V when compared to competing methods, as well as its competitive computational complexity for relatively small images.

%

The paper is organized as follows. Section~\ref{sec:back_notation} briefly reviews important background on linear mixing models and definitions and notation used for tensors. Section~\ref{sec_ProposedSolution} presents the proposed solution and the strategy to estimate tensor ranks. Section~\ref{sec:Simulations} presents the simulation results and comparisons. Finally, Section~\ref{sec:conclusions} presents the conclusions.

\section{Background and notation}
\label{sec:back_notation}
\subsection{Extended Linear Mixing Models}\label{sec:LMMs}

The Linear Mixing Model (LMM)~\cite{Keshava:2002p5667} assumes that a given pixel $\br_n = [r_{n,1},\,\ldots, \,r_{n,L} ]^\top$, with $L$ bands, is represented as
\begin{equation}
\begin{split}
    \br_n = \bM \balpha_n + \be_n,\,\,\,
    \text{subject to }\,\cb{1}^\top\balpha_n = 1 \text{ and } \balpha_n \succeq \cb{0}
\end{split}
\label{eq:LMM}
\end{equation}
where $\bM = [\bm_1,\,\ldots, \,\bm_R]$ is an $L\times R$ matrix whose columns are the $R$ endmembers $\bm_i = [m_{i,1},\,\ldots,\,m_{i,L}]^\top$, $\balpha_n = [\alpha_{n,1},\,\ldots,\,\alpha_{n,R}]^\top$ is the abundance vector, $\be\sim\mathcal{N}(0, \sigma_n^2\bI_L)$ is an additive white Gaussian noise (WGN), $\bI_L$ is the $L\times L$ identity matrix, and $\succeq$ is the entrywise $\geq$ operator. 
The LMM assumes that the pure material endmembers are fixed for all pixels $\br_n$, $n=1,\ldots,N$, in the HI. This assumption can jeopardize the accuracy of estimated abundances in many circumstances due to the spectral variability existing in a typical scene.

Different extensions of the LMM have been recently proposed to mitigate this limitation. These models employ a different endmember matrix for each pixel, and are particular cases of the model
\begin{align}
	\br_{n} {}={} \bM_{n}\balpha_{n} + \be_{n}
\end{align}
where $\bM_{n}$ is the endmember matrix for the $n$-th pixel. Different parametric models propose different forms for $\bM_n$ to account for spectral variability. These include additive perturbations over a mean matrix in the PLMM~\cite{Thouvenin_IEEE_TSP_2016}, multiplicative factors applied individually to each endmember in the ELMM~\cite{drumetz2016blind} or to each band in the GLMM~\cite{Imbiriba_glmm_2018}.
Moreover, spatial regularization of the multiplicative scaling factors in the ELMM and GLMM help to further mitigate the ill-posedness of the problem. Note that, although some works proposed to handle complex spectral variations indirectly by means of additive residual terms~\cite{Halimi_IEEE_Trans_CI_2017,hong2019augmentedLMMvariability}, these strategies usually do not estimate the EM spectra for each image pixel and, like the other models, also require carefully designed regularization strategies.

Other approaches pursue different ways to improve the conditioning of the inverse problem, employing for instance multiscale regularization on the abundance maps~\cite{borsoi2018superpix2}, or using additional information in the form of spectral libraries known a priori~\cite{uezato2018SU_variabilityAdaptiveBundlesDoubleSparse,drumetz2019SU_bundlesGroupSparsityMixedNorms} or extracted from the observed HI~\cite{borsoi2019deepGun}.

All these models, however, fail to exploit the high dimensional structure of the problem, which naturally suggests the representation of the HI, abundance maps, and endmember matrices for all pixels as higher-order tensors. In this work, instead of introducing a rigid parametric model for the endmembers, we employ a more general tensor model, using a well-devised low-rank constraint to introduce regularity to the estimated endmember tensor.

\subsection{Notation}
An order-$P$ tensor $\tensor{T} \in \amsmathbb{R}^{N_1\times \dots \times N_P}$ ($P>2$) is an $N_1\times \dots \times N_P$ array  with elements indexed by~$\tensor{T}_{n_1,n_2,\ldots,n_P}$. The $P$ dimensions of a tensor are called \textit{modes}. A mode-$\ell$ fiber of tensor $\tensor{T}$ is the one-dimensional subset of  $\tensor{T}$ obtained by fixing all but the $\ell$-th dimension, and is indexed by $\tensor{T}_{n_1,\ldots,n_{\ell-1},:,n_{\ell+1},\ldots,n_P}$. A slab or slice of tensor $\tensor{T}$ is a two-dimensional subset of $\tensor{T}$ obtained by fixing all but two of its modes.
An HI is often conceived as a three dimensional data cube, and can be naturally represented by an order-3 tensor $\tensor{R}\in\amsmathbb{R}^{N_1\times N_2\times L}$, containing $N_1\times N_2$ pixels represented by the tensor fibers $\tensor{R}_{n_1,n_2,:}\in\amsmathbb{R}^{L}$. 
Analogously, the abundances can also be collected in an order-3 tensor $\tensor{A}\in\amsmathbb{R}^{N_1\times N_2\times R}$. Thus, given a pixel $\tensor{R}_{n_1,n_2,:}$, the respective abundance vector $\balpha_{n_1,n_2}$ is represented by the mode-3 fiber $\tensor{A}_{n_1,n_2,:}$.
Similarly, the endmember matrices for each pixel can be represented as an order-4 tensor $\tensor{M}\in\amsmathbb{R}^{N_1\times N_2\times L \times R}$, where $\tensor{M}_{n_1,n_2,:,:}=\bM_{n_1,n_2}$.
We now review some operations of multilinear algebra (the algebra of tensors) that will be used in the following sections (more details can be found in~\cite{cichocki2015tensor}).


\subsection{Tensor product definitions}
\begin{definition}
\textbf{Outer product:}
The outer product between vectors $\cb{b}^{(1)}\in\amsmathbb{R}^{N_1},\cb{b}^{(2)}\in\amsmathbb{R}^{N_2},\ldots,\cb{b}^{(P)}\in\amsmathbb{R}^{N_P}$ is defined as the order-$P$ tensor $\tensor{T}=\cb{b}^{(1)}\circ\cb{b}^{(2)}\circ\cdots\circ\cb{b}^{(P)}\in\amsmathbb{R}^{N_1\times N_2\times\cdots\times N_P}$, where $\tensor{T}_{n_1,n_2,\ldots,n_P}=b^{(1)}_{n_1}b^{(2)}_{n_2}\cdots b^{(P)}_{n_P}$ and $b^{(i)}_{n_i}$ is the $n_i$-th position of $\cb{b}^{(i)}$. It generalizes the outer product between two vectors.
\end{definition}


\begin{definition}
\noindent\textbf{Mode-$k$ product:}
The mode-$k$ product, denoted $\tensor{U}=\tensor{T}\times_k\cb{B}$, of a tensor $\tensor{T} \in \amsmathbb{R}^{N_1\times \dots\times N_k\times \dots \times N_P}$ and a matrix $\cb{B} \in \amsmathbb{R}^{M_k\times N_k}$ is evaluated such that each mode-$k$ fiber of $\tensor{T}$ is multiplied by matrix $\cb{B}$, yielding $\tensor{U}_{n_1,n_2,\ldots,m_k,\ldots,n_P}=\sum_{i=1}^{N_k}\tensor{T}_{\ldots,n_{n-1},i,n_{n+1},\ldots} \cb{B}_{m_k,i}$.

\end{definition}

\begin{definition}
\noindent\textbf{Multilinear product:}
The full multilinear product, denoted by $\big\ldbrack\tensor{T};\bB^{(1)},\bB^{(2)},\ldots,\bB^{(P)}\big\rdbrack$, consists of the successive application of mode-$k$ products between $\tensor{T}$ and matrices $\bB^{(i)}$, represented as $\tensor{T}\times_1\bB^{(1)}\times_2\bB^{(2)}\times_3\ldots\times_P\bB^{(P)}$.
\end{definition}

\begin{definition}
\noindent\textbf{Mode-(M,1) contracted product:}
The contracted mode-M product, denoted by $\tensor{U}=\tensor{T}\times^M\cb{b}$, is a product between a tensor $\tensor{T}$ and a vector $\cb{b}$ in mode-M, where the resulting singleton dimension is removed, given by $\tensor{U}_{\ldots,n_{n-1},n_{n+1},\ldots}=\sum_{i=1}^{N_n}\tensor{T}_{\ldots,n_{n-1},i,n_{n+1},\ldots}\cb{b}_i$.
\end{definition}

\subsection{The Canonical Polyadic Decomposition}
\begin{figure}
\centering
\includegraphics[width=\linewidth]{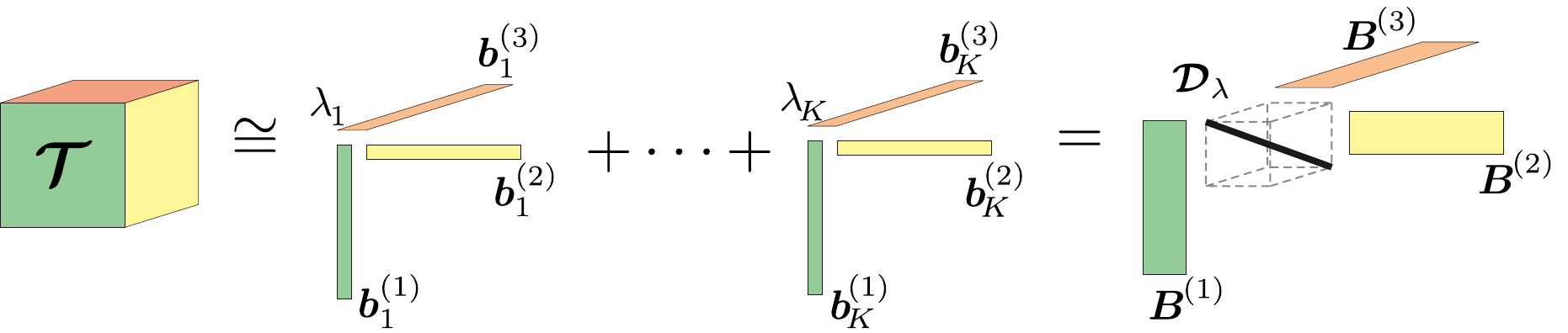}
\caption{Polyadic decomposition of a three-dimensional tensor, written as both outer products and mode$-n$ products.}
\label{fig:tensor_illustr}
\end{figure}

An order-$P$ rank-1 tensor is obtained as the outer product of $P$ vectors. The rank of an order-$P$ tensor $\tensor{T}$ is defined as the minimum number of order-$P$ rank-1 tensors that must be added to obtain $\tensor{T}$~\cite{sidiropoulos2017tensor}. Thus, any tensor $\tensor{T}\in\amsmathbb{R}^{N_1\times N_2\times\cdots\times N_P}$ with $\text{rank}(\tensor{T})=K$ can be decomposed as a linear combination of at least $K$ outer products of $P$ rank-1 tensors. This so-called \textit{polyadic decomposition} is illustrated in Figure~\ref{fig:tensor_illustr}. When this decomposition involves exactly $K$ terms, it is called the canonical polyadic decomposition  (CPD)~\cite{cichocki2015tensor} of a rank-$K$ tensor $\tensor{T}$, and is given by

%
%
\begin{equation}
\tensor{T} = \sum_{i=1}^{K} \lambda_i\cb{b}^{(1)}_i\circ\cb{b}^{(2)}_i\circ\cdots\circ \cb{b}^{(P)}_i.
\end{equation}
It has been shown that this decomposition is essentially unique under mild conditions~\cite{sidiropoulos2017tensor}. 
The CPD can be written alternatively using mode-$k$ products as
\begin{equation}
\tensor{T} = \tensor{D}_\lambda\times_1 \bB^{(1)}\times_2 \bB^{(2)}\cdots \times_{\!P} \bB^{(P)}
\end{equation}
or using the  full multilinear product as
\begin{equation}
\tensor{T} = \big\ldbrack \tensor{D}_\lambda;\bB^{(1)},\bB^{(2)},\ldots,\bB^{(P)} \big\rdbrack
\end{equation}
where $\tensor{D}_\lambda = \text{Diag}_P\big(\lambda_{1},\ldots,\lambda_{K}\big)$ is the $P$-dimensional diagonal tensor and $\bB^{(p)}=[\cb{b}^{(p)}_1,\ldots,\cb{b}^{(p)}_K]$, for $p=1,\ldots,P$.
%
%
Given a tensor $\tensor{T}\in\amsmathbb{R}^{N_1\times N_2\times\cdots\times N_P}$, the CPD can be obtained as the solution to the following optimization problem~\cite{sidiropoulos2017tensor}
\begin{equation}
\begin{split}
 \mathop{\min}_{\tensor{D}_\lambda,\bB^{(1)},\ldots,\bB^{(K)}} \,\,\,
 \frac{1}{2}\Big\|\tensor{T} - \sum_{i=1}^{K} \lambda_{i}\cb{b}_i^{(1)}\circ\cdots\circ\cb{b}_i^{(P)}\Big\|^2_F 
 .
\end{split}
\label{eq:cpd_opt_i}
\end{equation}

A widely used strategy to compute an approximate solution to~\eqref{eq:cpd_opt_i} is to use an alternating least-squares technique~\cite{sidiropoulos2017tensor}, which optimizes the cost function with respect to one term at a time, while keeping the others fixed, until convergence. Although optimization problem~\eqref{eq:cpd_opt_i} is generally non-convex, its solution is unique under relatively mild conditions, which is an important advantage of tensor-based methods~\cite{sidiropoulos2017tensor}.


\subsection{Tensor rank bounds}\label{sec:rank_bounds}
Finding the rank of an arbitrary tensor $\tensor{T}$ is NP-hard~\cite{haastad1990tensor}. 
In~\cite{sidiropoulos2017tensor}, upper and lower bounds on tensor ranks are presented for arbitrary tensors. Let $\tensor{T}$ be an order-3 tensor and
\begin{align}
R_1 \equiv & \dim \text{span} \{\tensor{T}_{:,j,k}\}_{\forall j,k}\nonumber \\
R_2 \equiv & \dim \text{span} \{\tensor{T}_{i,:,k}\}_{\forall i,k}\\
R_3 \equiv & \dim \text{span} \{\tensor{T}_{i,j,:}\}_{\forall i,j}\nonumber 
\end{align}
be the mode-1 (column), mode-2 (row) and mode-3 (fiber) ranks, respectively, of $\tensor{T}$.
Thus, the $K \equiv rank (\tensor{T})$ which is able to represent an arbitrary tensor is limited in the interval
\begin{equation}
 \max(R_1,R_2,R_3) \leq K \leq \min(R_1 R_2, R_1 R_3, R_2R_3).
\end{equation}
The reader can note that the bounds presented above often lead to very large tensor ranks. 
In many practical applications, however, the ``useful signal'' rank is often much less than the actual tensor rank~\cite{sidiropoulos2017tensor}.
Hence, when low-rank decompositions are employed to extract low-dimensional structures from the signal, the ranks that lead to meaningful results are usually much smaller than $\max(R_1,R_2,R_3)$.
In Section~\ref{sec:EstRank}, we propose a strategy to estimate the rank of tensor CPDs based on the variation of the multilinear singular values of $\tensor{T}$.


\section{Low-rank Unmixing Problem} \label{sec_ProposedSolution}

An effective strategy to capture the low-dimensional structures of HIs for solving the HU problem is to impose a low-rank structure to the abundance tensor~\cite{qian2017tensorNMFunmixing}. The same strategy can also be applied to the endmember tensor if one considers the endmember variabilities to be small or highly correlated in low-dimensional structures within the HI.
The low-rank property of HI tensors has been an important tool in the design of hyperspectral image completion~\cite{gandy2011tensorCompletionLowRankRecovery} and restoration algorithms~\cite{fan2017hyperspectralRestorationLowTensorRank}, consisting in one of the main low-dimensional structures that are currently being considered in hyperspectral imaging applications.
Thus, assuming that $\tensor{A}$ has a low-rank $K_\tensor{A}$, and that $\tensor{M}$ has a low-rank $K_\tensor{M}$ the global cost functional for the unmixing problem can be written as
%
%
\begin{align}\label{eq:unmixing_cost_func}
    J(\tensor{A},\tensor{M}) & {}={}
    \frac{1}{2}\!\sum_{n_1=1}^{N_1}\sum_{n_2=1}^{N_2}\|\tensor{R}_{n_1,n_2,:} -\tensor{M}_{n_1,n_2,:,:}\tensor{A}_{n_1,n_2,:} \|_F^2 
    \nonumber \\
    \text{s. t.}\;&\; \text{rank}(\tensor{M})= K_\tensor{M},\, \tensor{M}\succeq \cb{0}
    \\\nonumber
    &\; \text{rank}(\tensor{A})= K_\tensor{A},\,\tensor{A}\succeq \cb{0},\, \tensor{A}\times^3 \cb{1}_{R} = \cb{1}_{N_1\times N_2}.
\end{align}
Defining the HU problem as in~\eqref{eq:unmixing_cost_func} with fixed data independent ranks $K_\tensor{M}$ and $K_\tensor{A}$ limits its flexibility to adequately represent the desired abundance maps and endmember variability.
Though fixing low ranks for $\tensor{A}$ and $\tensor{M}$ tends to capture the most significant part of the tensors energy~\cite{Mei_lowrank_2018}, one may incur in a loss of fine and small scale details that may be relevant for specific data.
On the other hand, using large values for $K_{\tensor{A}}$ and $K_\tensor{M}$ makes the solution sensitive to noise, undermining the purpose of regularization. Thus, an important issue is how to effectively impose the low-rank constraint to achieve regularity in the solution without undermining its flexibility to adequately model small variations and details.

We propose to modify~\eqref{eq:unmixing_cost_func} by introducing new regularization terms, controlled by two low-rank tensors $\tensor{Q}\in\amsmathbb{R}^{N_1\times N_2\times R}$ and  $\tensor{P}\in\amsmathbb{R}^{N_1\times N_2\times L\times R}$, to impose non-strict constraints on $K_{\tensor{A}}$ and $K_{\tensor{M}}$. Doing that, tensors $\tensor{Q}$ and $\tensor{P}$ work as \emph{a priori} information, and the strictness of the low-rank constraint is controlled by two additional parameters $\lambda_\tensor{A},\, \lambda_\tensor{M}\in \amsmathbb{R}^+$. The proposed cost function is given by
%
%
%
\begin{equation}
\begin{split}
 J(\tensor{A},&\tensor{M}, \tensor{P},\tensor{Q}) =\\
  & \frac{1}{2}\sum_{n_1=1}^{N_1}\sum_{n_2=1}^{N_2}\|\tensor{R}_{n_1,n_2,:} -\tensor{M}_{n_1,n_2,:,:}\tensor{A}_{n_1,n_2,:} \|_F^2 \\
  & + \frac{\lambda_\tensor{M}}{2}\|\tensor{M} - \tensor{P}\|^2_F   + \frac{\lambda_\tensor{A}}{2}\|\tensor{A}-\tensor{Q}\|_F^2\\
  \text{s. t.}\;&\; \tensor{M}\succeq \cb{0},\; \tensor{A}\succeq \cb{0},\, \tensor{A}\times^{\!3} \cb{1}_{R} = \cb{1}_{N_1\times N_2}
\end{split}\label{eq:unmixing_cost_func2}
\end{equation}
with $\text{rank}(\tensor{P}) = K_\tensor{M}$ and $\text{rank}(\tensor{Q}) = K_\tensor{A}$. The optimization problem becomes
\begin{equation}
(\,\,\,\widehat{\!\!\!\tensor{A}},\,\widehat{\!\tensor{M}},\widehat{\tensor{P}},\widehat{\tensor{Q}}) = \mathop{\arg\min}_{\tensor{A},\tensor{M},\tensor{P},\tensor{Q}} J(\tensor{A},\tensor{M}, \tensor{P},\tensor{Q}).
\label{eq:opt_tensor}
\end{equation}
To solve~\eqref{eq:opt_tensor}, we propose to find a local stationary point by minimizing~\eqref{eq:unmixing_cost_func2} iteratively with respect to each variable. The resulting algorithm is termed the \textit{Unmixing with Low-rank Tensor Regularization Algorithm accounting for spectral Variability} (ULTRA-V), and is presented in $\text{Algorithm}$~1. The intermediate steps are detailed in the following.

\begin{algorithm} [bth]
\small
\SetKwInOut{Input}{Input}
\SetKwInOut{Output}{Output}
\caption{Global algorithm for solving \eqref{eq:unmixing_cost_func2}~\label{alg:global_opt2}}
\Input{$\tensor{R}$, $\lambda_\tensor{M}$, $\lambda_\tensor{A}$, $\tensor{A}^{(0)}$, and $\tensor{M}^{(0)}$.}
\Output{$\,\,\,\widehat{\!\!\!\tensor{A}}$ and $\,\widehat{\!\tensor{M}}$.}
$K_\tensor{Q}$ = estimateTensorRank($\tensor{A}^{(0)}$)\;
$K_\tensor{P}$ = estimateTensorRank($\tensor{M}^{(0)}$)\;
Set $i=0$ \;
\While{stopping criterion is not satisfied}{
$i=i+1$ \;
$\tensor{P}^{(i)} = \underset{\tensor{P}}{\arg\min} \,\,\,\,  {J}(\tensor{A}^{(i-1)},\tensor{M}^{(i-1)},\tensor{P})$ \;
$\tensor{Q}^{(i)} = \underset{\tensor{P}}{\arg\min} \,\,\,\,  {J}(\tensor{A}^{(i-1)},\tensor{M}^{(i-1)},\tensor{Q})$ \;
$\tensor{M}^{(i)} = \underset{\tensor{M}}{\arg\min} \,\,\,\,  {J}(\tensor{A}^{(i-1)},\tensor{M},\tensor{P}^{(i)},\tensor{Q}^{(i)})$ \;
$\tensor{A}^{(i)} = \underset{\tensor{A}}{\arg\min} \,\,\,\,  {J}(\tensor{A},\tensor{M}^{(i)},\tensor{P}^{(i)},\tensor{Q}^{(i)})$ \;
}
\KwRet $\,\,\,\widehat{\!\!\!\tensor{A}}=\tensor{A}^{(i)}$,~ $\,\widehat{\!\tensor{M}}=\tensor{M}^{(i)}$\;
\end{algorithm}

\subsection{Solving with respect to $\tensor{A}$}
To solve problem~\eqref{eq:opt_tensor} with respect to $\tensor{A}$ we use only the terms in~\eqref{eq:unmixing_cost_func2} that depend on $\tensor{A}$, leading to the cost function
\begin{equation}
\begin{split}
 J(\tensor{A}) =& \frac{1}{2}\sum_{n_1=1}^{N_1}\sum_{n_2=1}^{N_2}\|\tensor{R}_{n_1,n_2,:} -\tensor{M}_{n_1,n_2,:,:}\tensor{A}_{n_1,n_2,:} \|_F^2 \\
  & + \frac{\lambda_\tensor{A}}{2}\|\tensor{A} - \tensor{Q}\|^2_F \\
  \text{s. t.}\;&\; \tensor{A}\succeq \cb{0},\, \tensor{A}\times^3 \cb{1}_{R} = \cb{1}_{N_1\times N_2}
\end{split}
\end{equation}
which results in a standard regularized fully constrained least-squares problem that can be solved efficiently.

\subsection{Solving with respect to $\tensor{M}$}
Analogously to the previous section, to solve problem~\eqref{eq:opt_tensor} with respect to $\tensor{M}$, we use only the terms in~\eqref{eq:unmixing_cost_func2} that depend on $\tensor{M}$, leading to
\begin{align} \label{eq:opt_M_i}
 J(\tensor{M}) =& \frac{1}{2}\sum_{n_1=1}^{N_1}\sum_{n_2=1}^{N_2}\|\tensor{R}_{n_1,n_2,:} -\tensor{M}_{n_1,n_2,:,:}\tensor{A}_{n_1,n_2,:} \|_F^2 
 \nonumber\\
  & + \frac{\lambda_\tensor{M}}{2}\|\tensor{M} - \tensor{P}\|^2_F 
  \\ \nonumber
  \text{s. t.}\;&\; \tensor{M}\succeq \cb{0}.
\end{align}
which results in a regularized nonnegative least-squares problem.
An approximate solution can be obtained ignoring the positivity constraint over the endmember tensor and projecting the least-squares result onto the positive orthant as~\cite{Imbiriba_glmm_2018}
%
\begin{equation}
\begin{split}
 \,\widehat{\!\tensor{M}}_{n1,n2,:,:} = &\amsmathbb{P}_+\bigg(\Big(\tensor{R}_{n1,n2,:} \tensor{A}_{n1,n2,:}^\top + \lambda_\tensor{M}\tensor{P}_{n1,n2,:,:} \Big)
 \\
 &\quad\quad\quad\; \Big(\tensor{A}_{n1,n2,:} \tensor{A}^{\top}_{n1,n2,:}  +  \lambda_\tensor{M}\cb{I} \Big)^{-1}\bigg)
\end{split}
\end{equation}
where $\amsmathbb{P}_+:\amsmathbb{R}^{N_1\times N_1\times L}\rightarrow\amsmathbb{R}^{N_1\times N_1\times L}_+$ is the projection operator that maps every negative element to zero. Although this solution is approximate, it is significantly faster than directly solving~\eqref{eq:opt_M_i} and the algorithm still demonstrated good empirical convergence in our experiments.

\subsection{Solving with respect to $\tensor{P}$}
Rewriting the terms in~\eqref{eq:unmixing_cost_func2} that depend on $\tensor{P}$ leads to
\begin{equation}
 J(\bbLambda) = \frac{\lambda_\tensor{M}}{2}\|\tensor{M} - \tensor{P}\|^2_F.
 \label{eq:M_problem}
\end{equation}

Assuming that most of the energy of $\tensor{M}$ lies in a low-rank structure, we write the tensor $\tensor{P}$ as a sum of a small number $K_\tensor{P}$ of rank-1 components, such that
\begin{equation}
 \bbLambda = \sum_{i=1}^{K_\tensor{P}} \delta_i \bx_i^{(1)}\circ\bx_i^{(2)}\circ\bx_i^{(3)}\circ\bx_i^{(4)}.
 \label{eq:P_cpd}
\end{equation}
This introduces a low-rank \emph{a priori} condition on $\tensor{P}$ whose strictness can be controlled by the regularization constant $\lambda_\tensor{P}$.
Using~\eqref{eq:P_cpd} in~\eqref{eq:M_problem} leads to the  optimization problem
\begin{align}
 &\Big(\widehat{\cb{\Delta}},\widehat{\bX}^{(1)},\widehat{\bX}^{(2)},\widehat{\bX}^{(3)},\widehat{\bX}^{(4)}\Big) = \label{eq:P_als}\\ 
 &\!\! \mathop{\arg\min}_{\cb{\Delta},\bX^{(1)},\bX^{(2)},\bX^{(3)},\bX^{(4)}}\!\!\!\!\frac{\lambda_\tensor{M}}{2}\left\|\tensor{M} -\! \sum_{i=1}^{K_\tensor{P}} \! \delta_i \bx_i^{(1)}\!\circ\bx_i^{(2)}\!\circ\bx_i^{(3)}\!\circ\bx_i^{(4)}\right\|^2_F\nonumber
 \end{align}
where $\cb{\Delta} = \text{Diag}_4\left(\delta_{1}, \ldots, \delta_{K_\tensor{P}}\right)$ is a 4 dimensional diagonal tensor with $\cb{\Delta}_{i,i,i,i} = \delta_i$.
Problem~\eqref{eq:P_als} can be solved using an alternating least-squares strategy~\cite{sidiropoulos2017tensor}. 

Finally, the solution $\widehat{\tensor{P}}$ is obtained from $\widehat{\cb{\Delta}},\widehat{\bX}^{(1)},\widehat{\bX}^{(2)}$, $\widehat{\bX}^{(3)}$, and $\widehat{\bX}^{(4)}$  using the full multilinear product as
\begin{equation}
 \widehat{\tensor{P}} = \big\ldbrack \widehat{\cb{\Delta}};\widehat{\bX}^{(1)},\widehat{\bX}^{(2)},\widehat{\bX}^{(3)},\widehat{\bX}^{(4)} \big\rdbrack. 
\end{equation}

\subsection{Solving with respect to $\tensor{Q}$}
Analogous to the previous section, the cost function to be optimized for $\tensor{Q}$ can be written as
\begin{equation}
 J(\tensor{Q}) = \frac{\lambda_\tensor{A}}{2}\|\tensor{A} - \tensor{Q}\|^2_F .
 \label{eq:Q_problem}
\end{equation}

Assuming that most of the energy of $\tensor{A}$ lies in a low-rank structure, we write tensor $\tensor{Q}$ as a sum of a small number $K_\tensor{Q}$ of rank-1 components, such that
\begin{equation}
 	\tensor{Q} = \sum_{i=1}^{K_\tensor{Q}} 		
    \xi_{i}\bz^{(1)}_i\circ\bz^{(2)}_{i}\circ\bz^{(3)}_{i}.
 \label{eq:Q_cpd}
\end{equation}
This introduces a low-rank \emph{a priori} condition on $\tensor{A}$, which will be more or less enforced depending on the regularization constant $\lambda_\tensor{A}$. 
Using~\eqref{eq:Q_cpd} in~\eqref{eq:Q_problem} leads to the  optimization problem
\begin{align} 
 \Big(\widehat{\cb{\Xi}}\,\,,&\,\,\widehat{\bZ}^{(1)},\widehat{\bZ}^{(2)},\widehat{\bZ}^{(3)}\Big) =  \label{eq:Q_als}\\ 
 & \mathop{\arg\min}_{\cb{\Xi},\bZ^{(1)},\bZ^{(2)},\bZ^{(3)}}\frac{\lambda_\tensor{A}}{2}\Big\|\tensor{A} - \sum_{i=1}^{K_\tensor{Q}} \xi_{i}\bz_i^{(1)}\circ\bz_i^{(2)}\circ\bz_i^{(3)}\Big\|^2_F\nonumber
 \end{align}
where $\cb{\Xi} = \text{Diag}_3\big(\xi_{1}, \ldots, \xi_{K_{\!\tensor{Q}}}\big)$ is an order-3 diagonal tensor with $\cb{\Xi}_{i,i,i}=\xi_i$.
Problem~\eqref{eq:Q_als} can be solved using an alternating least-squares strategy~\cite{sidiropoulos2017tensor}.
Finally, the solution $\widehat{\tensor{Q}}$ is obtained from $\widehat{\cb{\Xi}},\widehat{\bZ}^{(1)},\widehat{\bZ}^{(2)}$ and $\widehat{\bZ}^{(3)}$ using the full multilinear product as
\begin{equation}
 \widehat{\tensor{Q}} = \big\ldbrack \widehat{\cb{\Xi}} \,\,;\, \widehat{\bZ}^{(1)}; \widehat{\bZ}^{(2)};\widehat{\bZ}^{(3)} \big\rdbrack.
\end{equation}

\subsection{Computational complexity of Algorithm~\ref{alg:global_opt2}}\label{sec:complexity}

The computational complexity of each iteration of Algorithm~\ref{alg:global_opt2} can be measured as follows. The optimizations w.r.t. $\tensor{A}$ and $\tensor{M}$ both consist of regularized constrained LS problems with $N_1N_2 R$ and $N_1N_2LR$ variables respectively. Thus, these problems can be solved with a complexity of $\mathscr{O}\big((N_1N_2 R)^3\big)$ and $\mathscr{O}\big((N_1N_2 L R)^3\big)$, respectively. The optimizations w.r.t. variables $\tensor{P}$ and $\tensor{Q}$ consist of CP decompositions of these tensors with ranks $K_{\tensor{P}}$ and $K_{\tensor{Q}}$, respectively. Considering an alternating least squares (ALS) approach for the CPD, these optimization problems will have computational complexities of $\mathscr{O}\big(K_{iter}K_{\tensor{P}}N_1N_2LR\big)$ and $\mathscr{O}\big(K_{iter}K_{\tensor{Q}}N_1N_2R\big)$, respectively, where $K_{iter}$ is the number of ALS iterations~\cite{sorber2013optimizationAlgsTensorDec}.
Thus, the overall complexity of the algorithm scales linearly with the number of ALS iterations and with the tensor ranks, and cubically in the problem dimensions. When processing large datasets, the extra complexity could be partially mitigated by applying image segmentation or band
selection~\cite{Imbiriba2017_bs_tip,wang2018spectral,wang2018optimal} strategies. This analysis is beyond the scope of the present work and will be addressed in the future.

\subsection{Estimating tensor ranks}\label{sec:EstRank}



In Section~\ref{sec:rank_bounds} we have recalled important results relating bounds for order-3 tensor ranks to the span of the  matricized versions of tensors. We have also noted, from our own experience, that those bounds tend to indicate tensor ranks that are larger than the rank associated with the information relevant for HU. Our interest in HU is to model low-dimensional structures of the HI using low-rank tensors. At the same time, this low-rank representation should be rich enough to include all dimensions of the original HI tensor that contain relevant information. 
Therefore, although the literature presents many rank estimation strategies (see~\cite{yokota2017robustTensorRankEstimation} and references therein), in this work we exploit the rank bounds discussed in Section~\ref{sec:rank_bounds} to approximate the ``useful rank'' of a tensor by the number of the largest singular values of their matricized versions required to represent most of the tensor energy.

Let $\cb{T}_i = \text{mat}_i(\tensor{T}) \in \amsmathbb{R}^{N_i\times (N_1\ldots N_{i-1} N_{i+1},\ldots N_P)}$ be the matricization of an arbitrary tensor $\tensor{T}\in \amsmathbb{R}^{N_1\times N_2\ldots\times N_P}$ obtained by stacking all tensor fibers along the $i$-th tensor dimension. Let $\cb{s}_i = \text{SVD}(\cb{T}_i)$ be the set of singular values of $\cb{T}_i$, sorted descending in value. Also, let $\cb{d}_i = \text{diff}(\cb{s}_i)$ be the vector of first order differences of the elements of $\cb{s}_i$, such that, $d^{(i)}_j = s^{(i)}_j-s^{(i)}_{j+1}$. Then, we define the $i$-th candidate for rank of $\tensor{T}$ as the smallest index $j$ such that $|d^{(i)}_j|$ sufficiently small, namely, 
\begin{equation}
\hat{R}_i =  \min j,\, \text{s.t.},\,  |d^{(i)}_j|< \varepsilon
\end{equation}
where $\varepsilon$ is a parameter limiting the singular value variation. In all experiments reported here we used $\varepsilon = 0.15$. We have experimentally verified that the resulting abundance MSE has very low sensitivity to the choice of $\varepsilon$.
Finally, we approximate the rank of tensor $\tensor{T}$ as

\begin{equation} \label{eq_RankDefinition}
 K =  \max \{\hat{R}_1,\ldots,\hat{R}_P\}. 
\end{equation}

For the experiments reported in this paper, we have used definition~\eqref{eq_RankDefinition} to estimate $K_\tensor{P}$ and $K_\tensor{Q}$ in~\eqref{eq:P_cpd} and~\eqref{eq:Q_cpd} from the abundance and endmember tensors estimated using simple unmixing strategies such as the scaled constrained least squares (SCLS)~\cite{drumetz2016blind}.

\begin{table}[htb!]
\caption{Simulation results using synthetic data.}
\centering
\renewcommand{\arraystretch}{1.2}
\begin{tabular}{lp{0.8cm}p{0.8cm}p{0.8cm}p{0.8cm}p{0.8cm}p{0.6cm}}
\toprule
\multicolumn{6}{c}{Data Cube 0 -- DC0} \\
\toprule\bottomrule
& $\text{MSE}_{\tensor{A}}$ & $\text{MSE}_{\tensor{M}}$ &$\text{SAM}_{\tensor{M}}$ & $\text{MSE}_{\tensor{R}}$ &$\text{SAM}_{\tensor{R}}$ & Time\\ \midrule
FCLS	&	1.81		&	-		&	-			&	16.11		&	5.92	&	0.42	\\
SCLS	&	0.68		&	175.79		&	6.19			&	59.38		&	5.42	&	0.38	\\
PLMM	&	0.76		&\bf{\cblue{94.57}}	&\bf{\cblue{5.58}}		&	4.89		&	3.41	&	81.89	\\
ELMM	&	0.35		&	106.17		&	5.63			&	4.78		&	3.40	&	17.15	\\
GLMM	&\bf{\cblue{0.34}}	&	101.51		&	5.87			&\bf{\cred{5.7e-3}}	&\bf{\cred{0.09}}&	20.23	\\
ULTRA-V	&\bf{\cred{0.23}}	&\bf{\cred{92.39}}		& \bf{\cred{5.56}}	&\bf{\cblue{0.73}}	&\bf{\cblue{1.32}}&	14.46	\\
\cred{ULTRA} & 1.81 & - & - & 16.11 & 5.92 & 1.42\\
\toprule										
\multicolumn{6}{c}{Data Cube 2 -- DC1}	\\
\toprule\bottomrule	
FCLS	&	2.01		&	-		&	-		&	6.93		&	3.71		&	0.74	\\
SCLS	&	2.07		&	92.16		&	\bf\cblue{5.37}		&	24.78		&	3.41		&	0.76	\\
PLMM	&	1.58		& 157.00	& 8.49	&	2.75		&	2.41		&	120.48	\\
ELMM	&	1.29		&	69.15		&	5.95		&\bf{\cblue{0.01}}	&\bf{\cblue{0.09}}	&	23.18	\\
GLMM	& 1.20	&	\bf{\cblue{68.11}}		&	6.09		&\bf{\cred{0.01}}	&\bf{\cred{0.08}}	&	29.81	\\
ULTRA-V &\bf{\cred{1.12}}	&\bf{\cred{60.12}}	&\bf{\cred{5.21}}	&	4.26		&	2.56		&	29.47	\\
\cred{ULTRA} & \bf{\cblue{1.17}}  & - & - & 10.93  & 3.94 & 4.44\\
\toprule											
\multicolumn{6}{c}{Data Cube 3 -- DC2}	\\						
\toprule\bottomrule									
FCLS	&	1.90		&	-		&	-		&	2.03		&	13.69		&	0.30	\\
SCLS	&	0.71		&\bf{\cred{1.66}}	&\bf{\cblue{2.29}}		&\bf{\cblue{1.07}}&	12.68		&	0.30	\\
PLMM	&	1.27		&	2.45		&\bf{\cred{2.28}}	&	2.12		&	12.17		&	61.13	\\
ELMM	&	0.63		&	2.84		&	3.32		&	2.31		&	13.13		&	11.15	\\
GLMM	&\bf{\cblue{0.59}}	&\bf{\cblue{1.84}}	&	2.79		&	2.04		&\bf{\cblue{12.58}}		&	11.60	\\
ULTRA-V	&\bf{\cred{0.46}}	&	2.91		&	2.94		&\bf{\cred{9e-5}}	&\bf{\cred{0.49}}	&	5.19	\\
\cred{ULTRA} & 0.83 & - & - & 1.02 & 3.73 & 0.36\\
\bottomrule\toprule
\end{tabular}
\label{tab:results_synthData}
\end{table}
 

\section{Simulations}\label{sec:Simulations}
In this section, the performance of the proposed methodology is illustrated through simulations with both synthetic and real data. We compare the proposed ULTRA-V method with the the fully constrained least squares (FCLS), the scaled constrained least squares (SCLS)~\cite{drumetz2016blind}, the PLMM~\cite{Thouvenin_IEEE_TSP_2016}, the ELMM~\cite{drumetz2016blind}, and the GLMM~\cite{Imbiriba_glmm_2018}. To highlight the differences between ULTRA-V and ULTRA~\cite{imbiriba2018_ULTRA}, we also consider ULTRA for simulations with synthetic data.



To measure the accuracy of the unmixing methods we consider the  Mean Squared Error (MSE)
\begin{equation}
\text{MSE}_{\tensor{X}} = {\frac{1}{N_{\tensor{X}}}\|\text{vec}(\tensor{X})- \text{vec}(\widehat{\tensor{X}})\|^2} 
\end{equation}
where $\text{vec}(\cdot)$ is the vectorization operator $\tensor{X}\rightarrow \bx$, $\amsmathbb{R}^{a\times b\times c}\mapsto \amsmathbb{R}^{abc}$, $N_{\tensor{X}} = abc$, and the Spectral Angle Mapper for the HI 



\begin{equation}
  \text{SAM}_{\tensor{R}} = \frac{1}{N}\sum_{n=1}^{N}\arccos\left(\frac{\br_n^\top\widehat{\br}_n}{\|\br_n\|\|\widehat{\br}_n\|}\right)
\end{equation}
and for the endmembers tensor
\begin{equation}
  \text{SAM}_{\tensor{M}} = \frac{1}{N}\sum_{n=1}^{N}\sum_{k=1}^{R}\arccos\left(\frac{\bm_{k,n}^\top\widehat{\bm}_{k,n}}{\|\bm_{k,n}\|\|\widehat{\bm}_{k,n}\|}\right).
\end{equation}

All the algorithms were implemented in Matlab on a desktop computer equipped with an Intel Core~I7 processor with 4.2Ghz and 16Gb of RAM.
In all cases, we used endmembers extracted using the VCA~\cite{Nascimento2005} either to build the reference endmember matrix or to initialize the different methods, with the number of endmembers~$R$ assumed to be known a priori. The abundance maps were initialized using the maps estimated by the~SCLS.

\begin{figure}[htb]
\centering
\includegraphics[width=\linewidth]{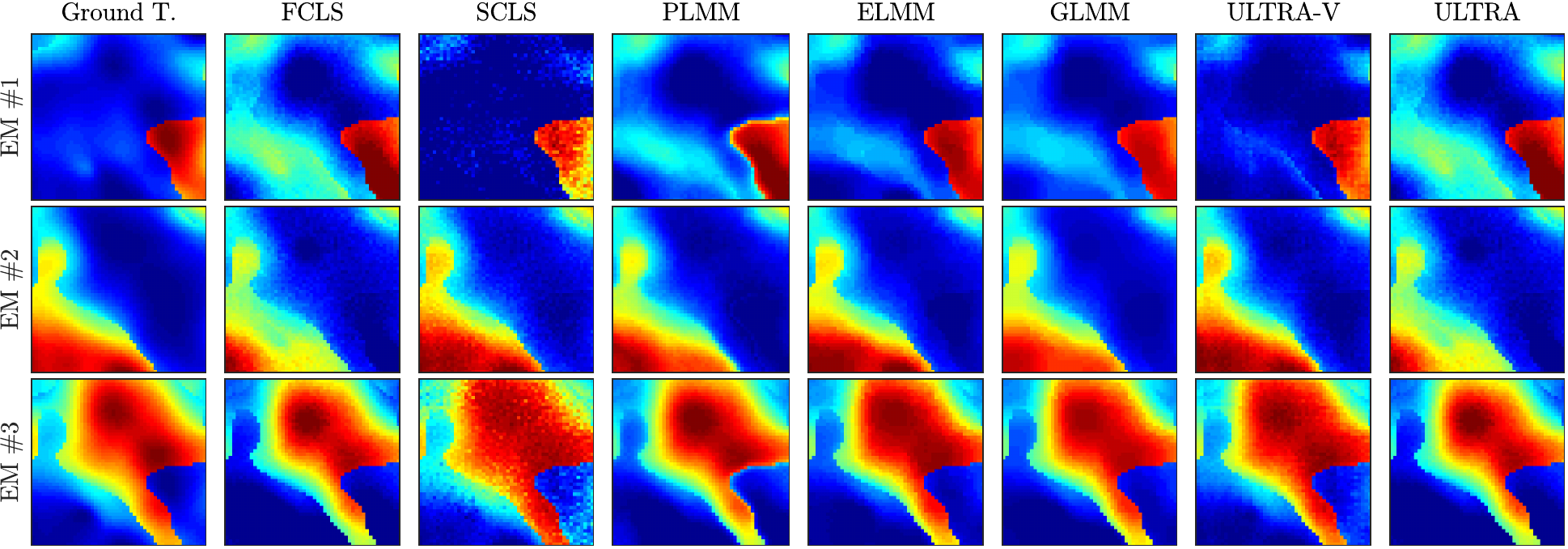}\\
\includegraphics[width=\linewidth]{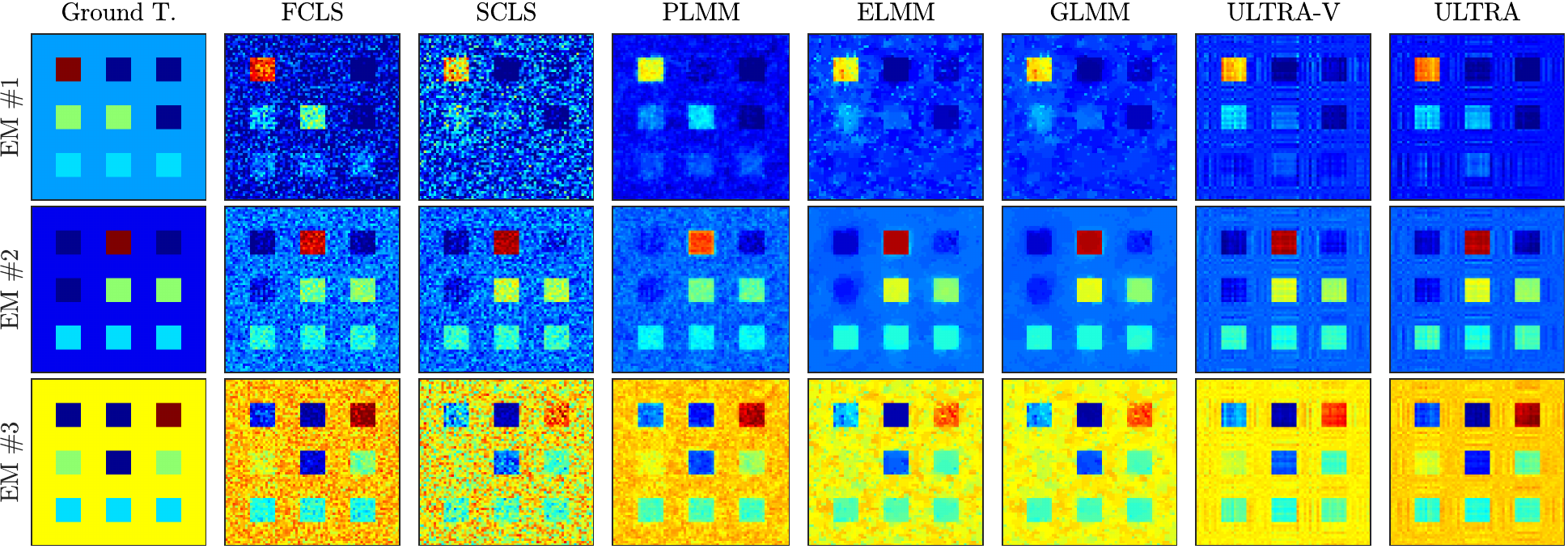}\\
\includegraphics[width=\linewidth]{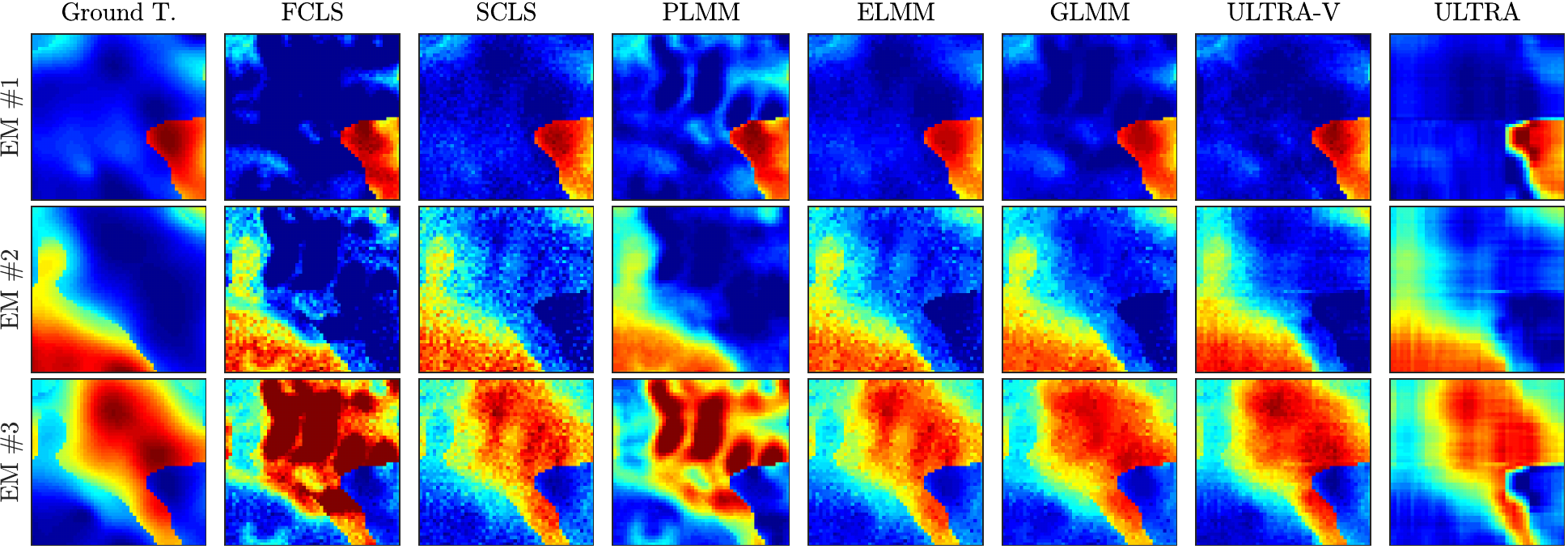}
\caption{Abundance maps of (top--down) DC0, DC1, and DC2  for all tested algorithms. Abundance values represented by colors ranging from blue ($\alpha_k = 0$) to red ($\alpha_k = 1$).}\label{fig:ab_maps_dc2}
\end{figure}

\subsection{Synthetic data}
For a comprehensive comparison among the different methods we created three synthetic datasets, namely Data Cube 0 (DC0),  Data Cube 1 (DC1) and Data Cube 2 (DC2), with 50$\times$50 pixels (DC0 and DC2) and 70x70 pixels (DC1). 
DC0 and DC1 were built using three 224-band endmembers extracted from the USGS Spectral Library~\cite{clark2003imaging}, while DC2 was built using three 16-band minerals often found in bodies of the Solar System. For the three datasets, spatially correlated abundance maps were used, as depicted in the first column of Fig.~\ref{fig:ab_maps_dc2}. 
For DC0, we adopted the variability model used in~\cite{drumetz2016blind} (a multiplicative factor acting on each endmember). For DC1, we used the variability model according to the PLMM~\cite{Thouvenin_IEEE_TSP_2016}. For DC2, we used the Hapke model~\cite{Hapke1981} devised to realistically represent the spectral variability introduced due to changes in the illumination conditions caused by the topography of the scene~\cite{drumetz2016blind}. White Gaussian noise was added to all datasets to yield a 30dB SNR.


To select the optimal parameters for each algorithm, we performed grid searches for each dataset. We used parameter search ranges based on the ranges tested and discussed by the authors in the original publications. For the PLMM we used $\gamma=1$, since the authors fixed this parameter in all simulations, and searched for $\alpha$ and $\beta$ in the ranges $[10^{-6},\, 10^{-3},\, 0.1,\, 0.35,\, 0.7,\, 1.4,\, 5,\, 25]$ and $[10^{-9},\, 10^{-5},\, 10^{-4},\, 10^{-3}]$, respectively. For both ELMM and GLMM, we selected the parameters among the following values: $\lambda_{S},\,\lambda_M \in [0.01,\, 0.1,\, 1,\, 5,\, 10,\, 15]$, $\lambda_{A} \in [10^{-6},\, 10^{-3},\, 0.01,\, 0.05,\, 0.1,\, 1,\, 10$, and $\lambda_\psi,\,\lambda_{\bbPsi} \in [10^{-6},\, 10^{-3},\, 10^{-1}]$, while for the proposed ULTRA-V we selected the parameters in the intervals $\lambda_\tensor{A}\in [0.001,\, 0.01,\, 0.1,\, 1,\, 10,\, 100]$ and $\lambda_\tensor{M} \in [0.1,\, 0.2,\, 0.4,\, 0.6,\, 0.8,\, 1]$. For the ULTRA we searched $\lambda_\tensor{A}$ in the same interval used for the ULTRA-V.

The results are shown in Table~\ref{tab:results_synthData}, were the best and second best results for each metric are marked in bold red and bold blue, respectively. ULTRA-V clearly outperformed the competing algorithms for all datasets in terms of $\text{MSE}_{\tensor{A}}$. For the other metrics, the best results depended on the datasets. 
In terms of $\text{MSE}_\tensor{M}$ and $\text{SAM}_\tensor{M}$, ULTRA-V yielded the best results for DC0 and DC1. Finally, ULTRA-V results for $\text{MSE}_\tensor{R}$ and $\text{SAM}_\tensor{R}$ were the second best for DC0 and the best for DC2. 
The execution times, shown in the rightmost columns of Table~\ref{tab:results_synthData}, show that ULTRA-V required the smallest execution time among the more sophisticated algorithms (PLMM, ELMM and GLMM) for DC0 and DC2, and comparable execution time for DC1.
As expected, the ULTRA method provided results that were often better than the FCLS but significantly worse than those obtained from methods accounting for EM variability. This happens because ULTRA imposes a low-rank structure over the abundances but keeps the EMs fixed for all pixels, what greatly limits the algorithm capacity to adapt to EM variations along the image.

\begin{figure}[htb]
\centering
\includegraphics[width=0.33\linewidth]{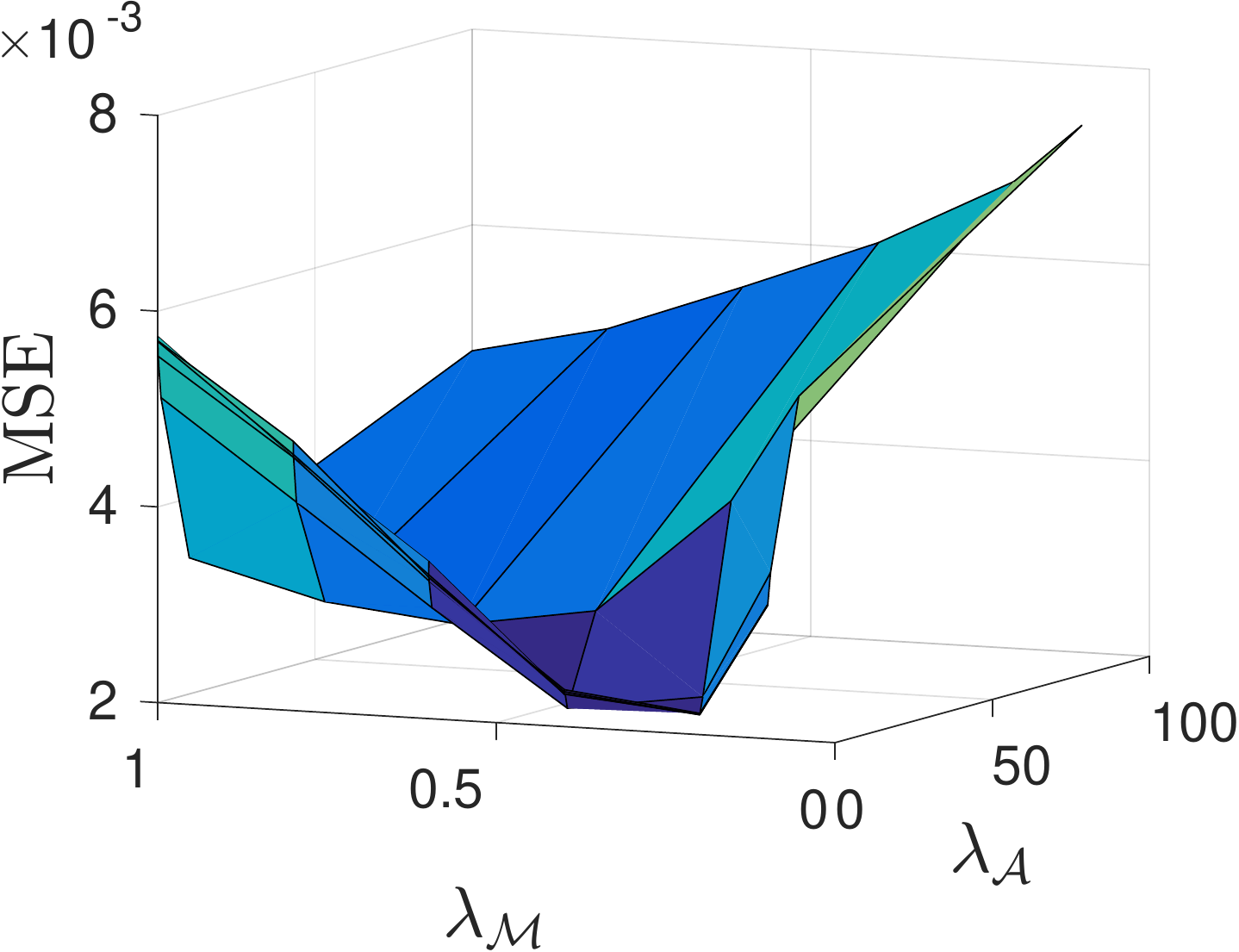} \hspace{-0.3cm}
\includegraphics[width=0.33\linewidth]{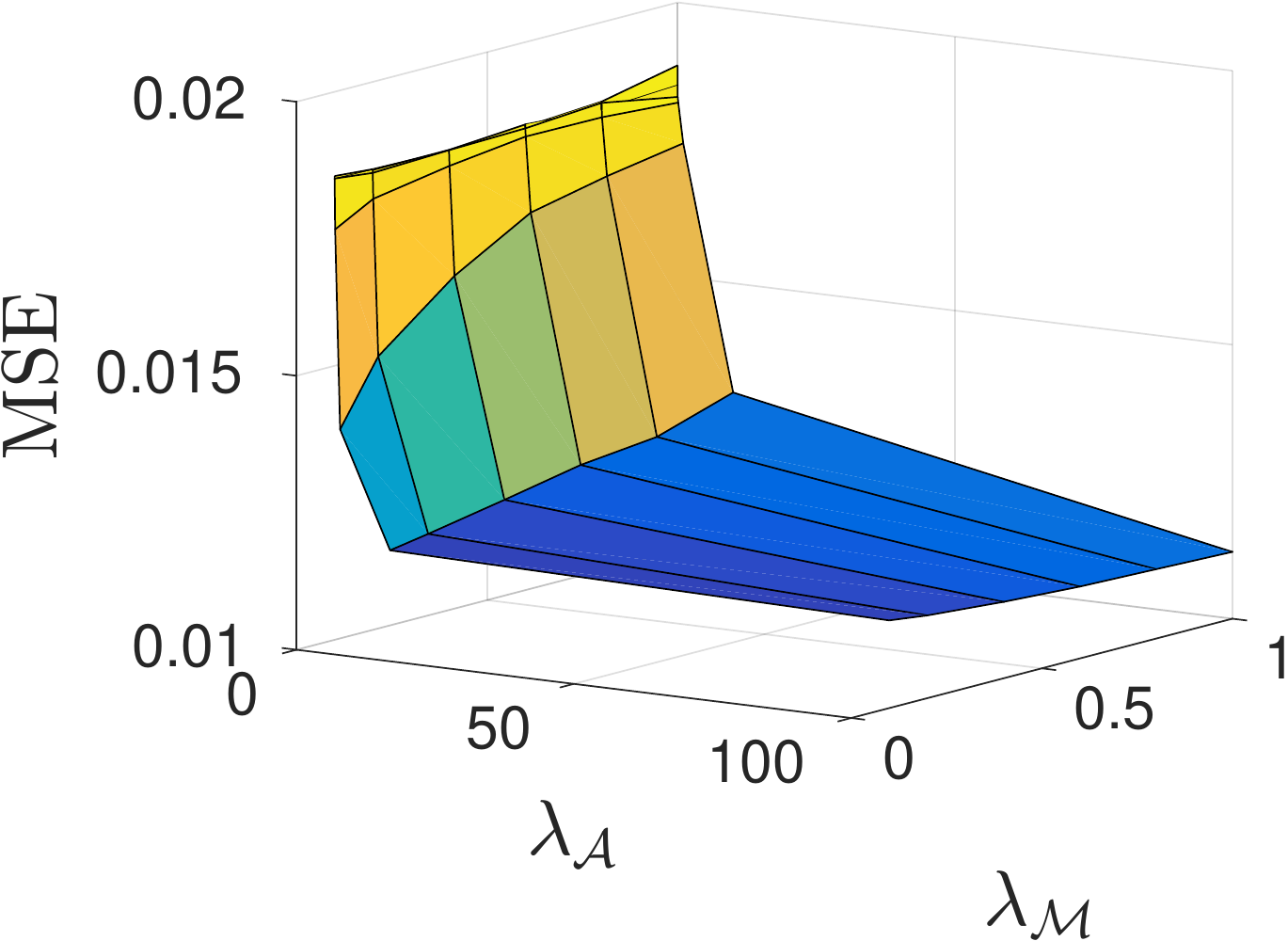} \hspace{-0.2cm}
\includegraphics[width=0.33\linewidth]{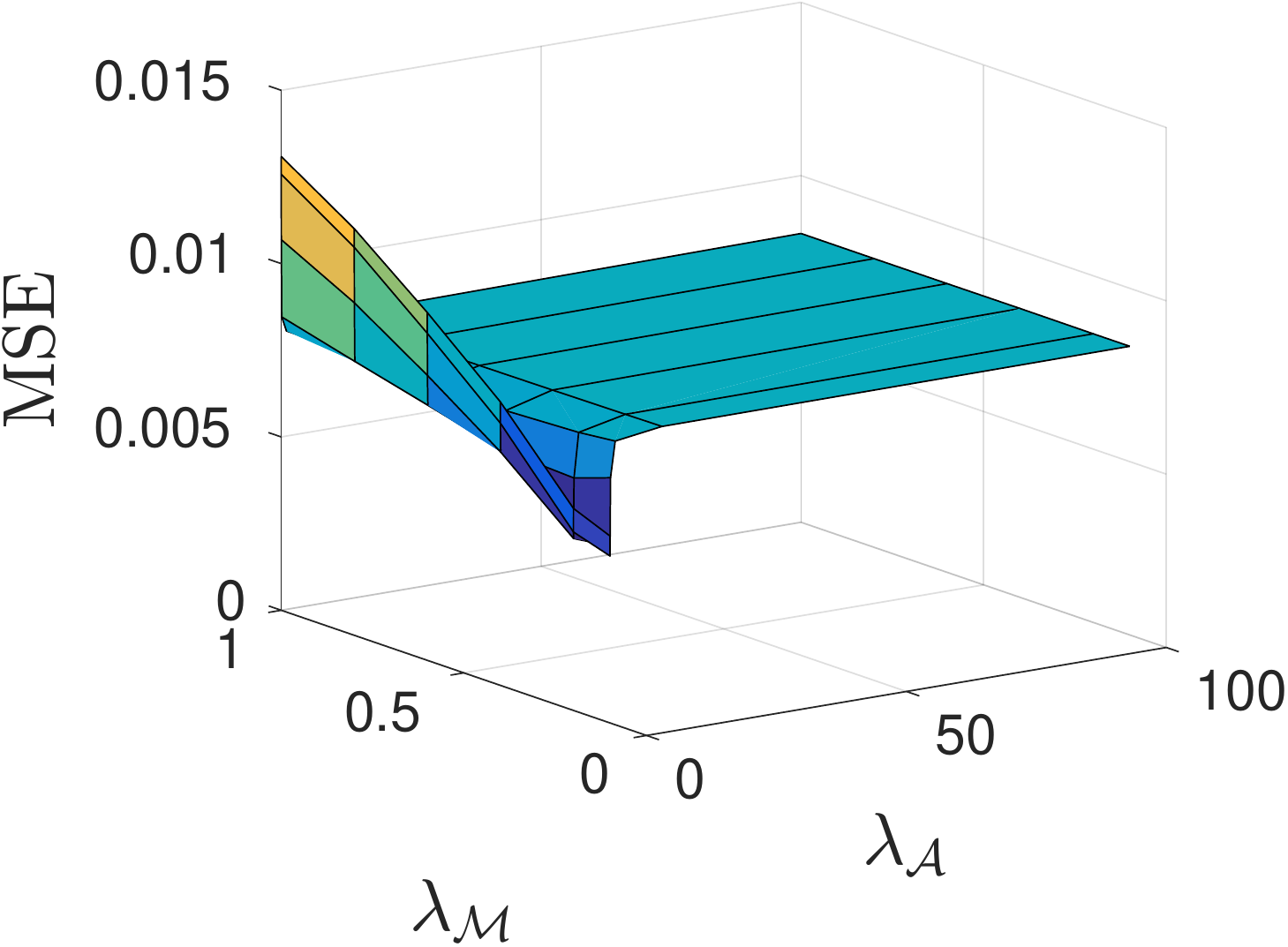}
\caption{Parameters sensitivity to changes around the optimal values. Left: DC0, Middle: DC1, and Right: DC2.}\label{fig:par_sen}
\end{figure}

\subsubsection{Parameters sensitivity}

While we have proposed a strategy to determine rank values for tensors $\tensor{P}$ and $\tensor{Q}$, the parameters $\lambda_\tensor{M}$ and $\lambda_\tensor{A}$ need to be selected by the user. 
We now study the sensitivity of the ULTRA-V performance to variations of the parameters within the parameter search intervals presented in the previous section. Figure~\ref{fig:par_sen} shows the values of $\text{MSE}_{\tensor{A}}$ resulting from unmixing the data using each combination of the parameter values.  The sensitivity clearly tends to increase when values less than 1 are used for both parameters. Our practical experience indicates that good $\text{MSE}_{\tensor{A}}$ results can be obtained using $\lambda_\tensor{M}$ in $[0,1]$, and large values about 100 for $\lambda_\tensor{A}$. Moreover, some insensitivity is verified for small changes in $\lambda_\tensor{A}$ about large values. Thus, searching $\lambda_\tensor{A}$ in $[0.001, 100]$ with values spaced by decades as done for the examples in the previous section seems reasonable.

\subsubsection{Discussion}

A close look at Fig.~\ref{fig:ab_maps_dc2} reveals the abundance maps estimated using ULTRA-V look noisier than those obtained using ELMM and GLMM. This is because the proposed approach does not impose the local smoothness imposed by total variation (TV), but emphasizes abundance regularity by enforcing a low-rank property.  We note, however, that the spacial smoothness imposed by TV is not necessarily mandatory for a good abundance estimation, as can be verified from the results in Table~\ref{tab:results_synthData}.


\begin{figure}[htb!]
\centering
\includegraphics[width=\linewidth]{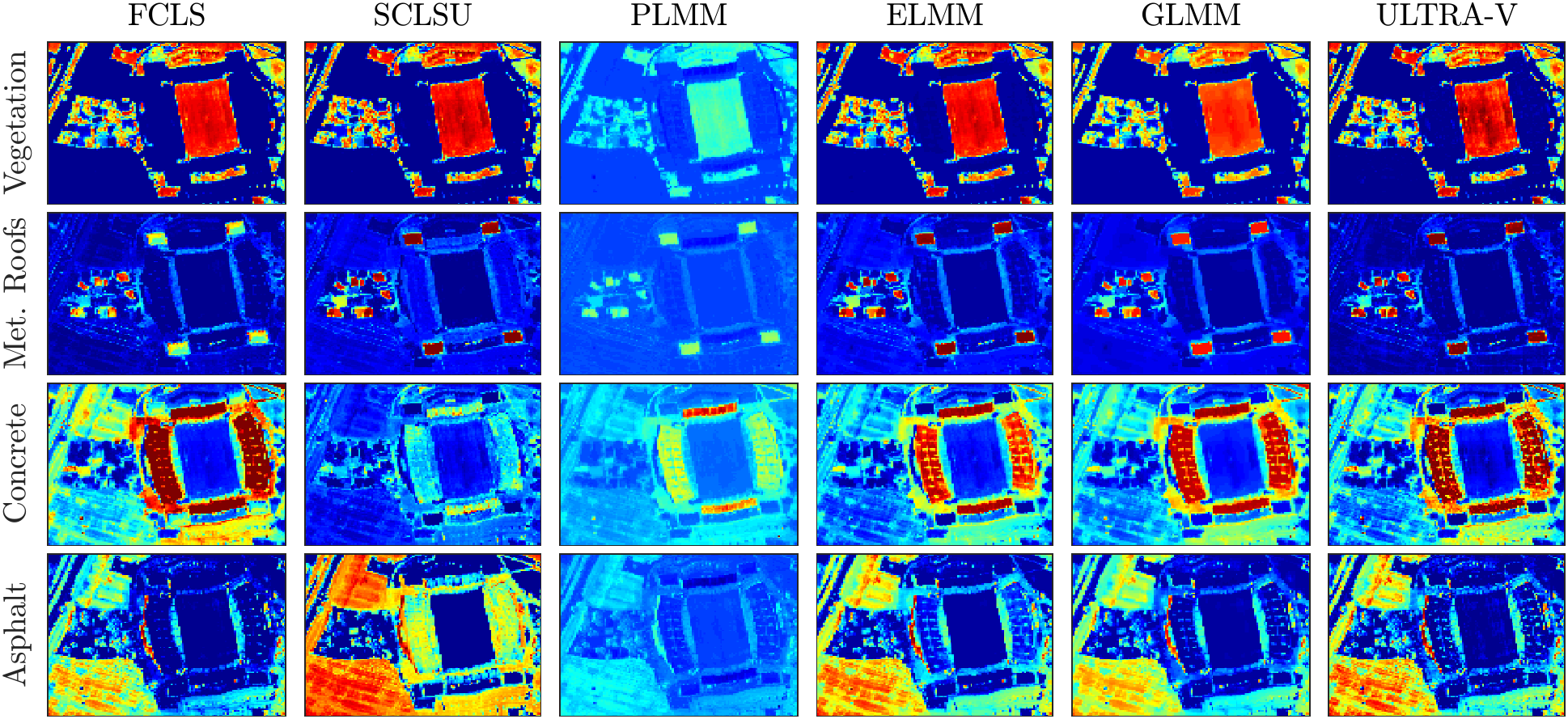}\\
 \includegraphics[width=\linewidth]{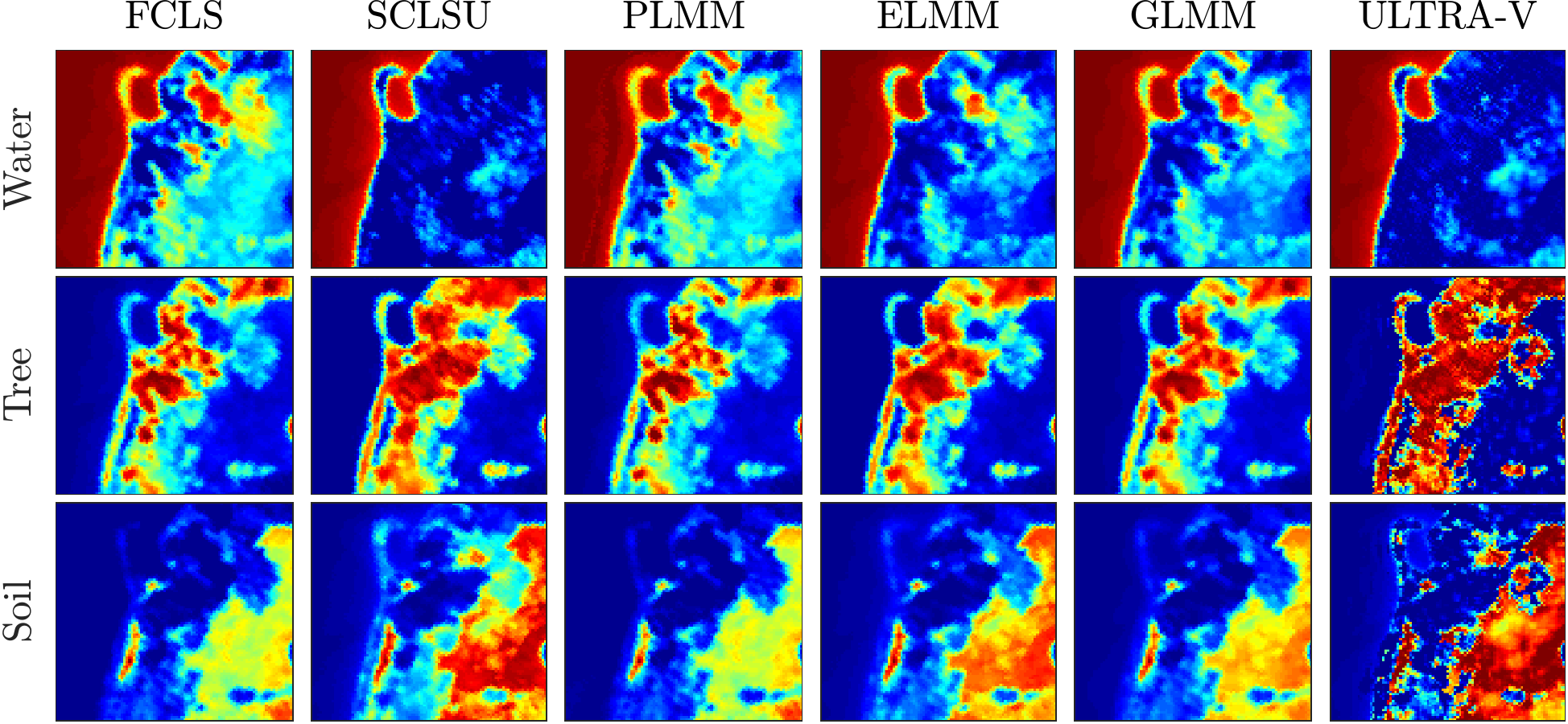}\\
 \includegraphics[width=\linewidth]{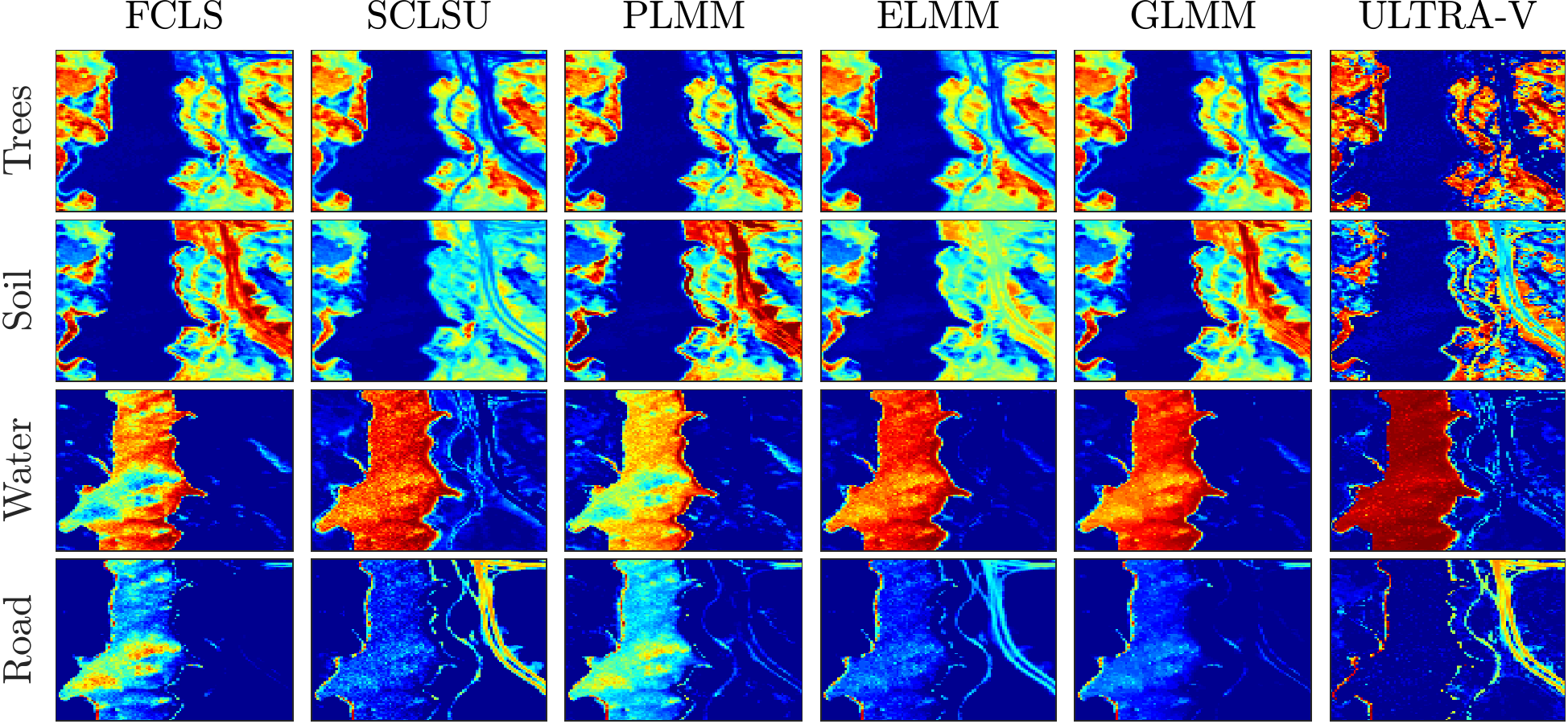}
\caption{Abundance maps of the Houston (upper panel),  Samson (middle painel), and Jasper Ridge (bottom painel) data sets for all tested algorithms. Abundance values represented by colors ranging from blue ($\alpha_k = 0$) to red ($\alpha_k = 1$).}\label{fig:ab_maps_realdata}
\end{figure}



\subsection{Real data}

For the simulations with real data, we considered three datasets, consisting of the Houston, Samson and Jasper Ridge images.
All datasets were captured by the AVIRIS, which originally has 224 spectral bands. For all images, the water absorption bands were removed resulting in 188 bands for the Houston image, 156 bands for the Samson image and 198 bands for the Jasper Ridge image.
The Houston data set is known to have four predominant endmembers~\cite{drumetz2016blind,borsoi2019tensorInterpolationICASSP}.  The Samson and Jasper Ridge images are known to have three and four endmembers, respectively~\cite{borsoi2018multiscale1}. For all images the endmembers were extracted using the VCA~\cite{Nascimento2005}.
Fig.~\ref{fig:ab_maps_realdata} shows the reconstructed abundance maps for all images and for all tested methods. 
The quantitative results are shown in Table~\ref{tab:results_realdata}. Note that since the ground truth (correct) abundance values are not available for these images, only the reconstruction error $\text{MSE}_{\tensor{R}}$ has been used as a sort of quality verification.

The last column in Fig.~\ref{fig:ab_maps_realdata} shows that the proposed ULTRA-V method provided an  accurate abundance estimation, clearly outperforming the competing algorithms\footnote{The differences between the ELMM and ULTRA-V results are less significant for the Houston image.}. In fact, for the Concrete and Metallic Roofs endmembers, the ULTRA-V abundance map presents stronger Concrete and Metallic Roofs components in the stadium stands and stadium towers, respectively, when compared with the other methods equipped for dealing with spectral variability. The performance improvement provided by ULTRA-V is clearer for the Samson and Jasper Ridge images. For instance, there is significantly less confusion between the Water, Tree and Soil endmembers in the ULTRA-V results for the Samson image when compared to those of the PLMM, ELMM, and GLMM methods. Similarly, the ULTRA-V reconstructed abundance maps of the Jasper Ridge image show a much stronger Water component in the river and less confusion between the Tree, Soil and Road endmembers.




The objective metrics presented in Table~\ref{tab:results_realdata} indicate that ULTRA-V yields competitive reconstruction errors in terms of MSE. These results, however, should be interpreted with proper care, as the connection of reconstruction error and abundance estimation is not straightforward. 

The execution times in Table~\ref{tab:results_realdata} indicate that, as discussed in Section~\ref{sec:complexity}, ULTRA-V did not scale well with the larger image sizes and higher number of endmembers, which directly impacted the CPD stage of ULTRA-V. Moreover, the more complex images resulted in higher rank estimates using the strategy discussed in Section~\ref{sec:EstRank}. This indicates that there is still room for improving the proposed method by either providing a segmentation strategy or using faster CPD methods. This, however, is an open problem that will be addressed in future works.

\begin{figure}[htb]
    \centering
    \includegraphics[width=\linewidth]{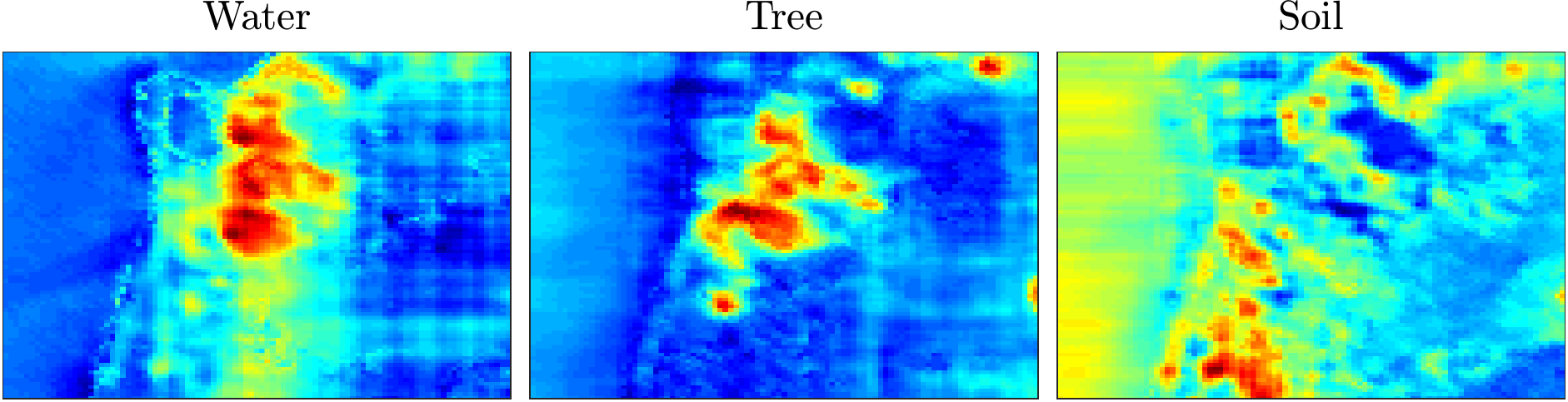}
    \caption{Average of the ULTRA-V endmembers tensor projection over the 3 principal components for the Samson data set.}
    \label{fig:samson_em_pca}
\end{figure}

\begin{figure}[htb]
    \centering
    \includegraphics[width=\linewidth]{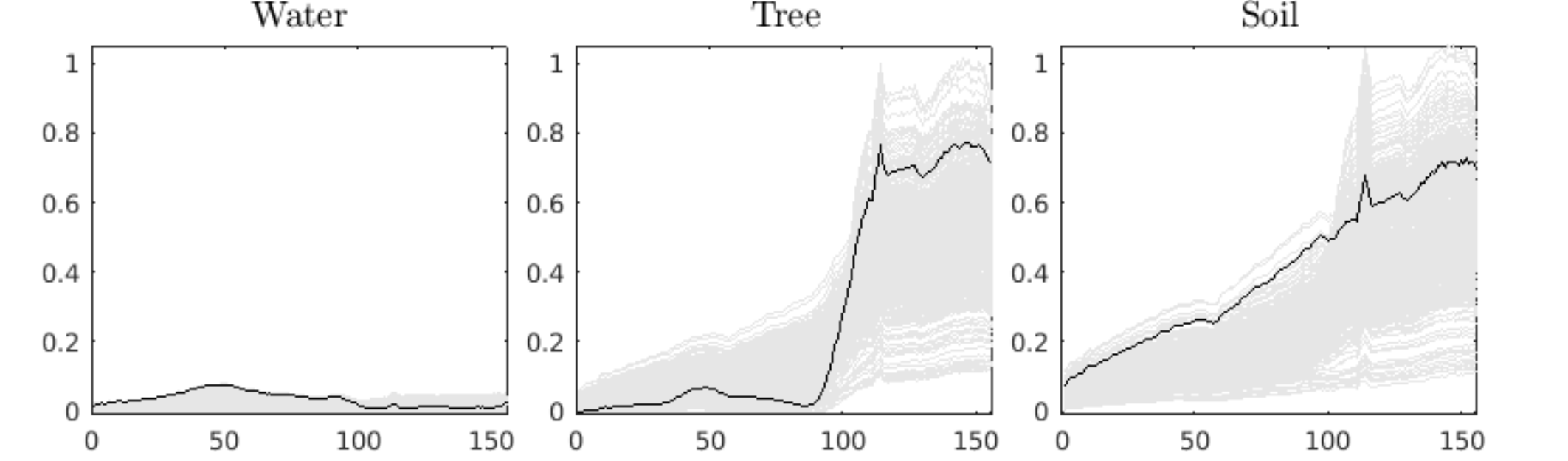}
    \caption{VCA result (black) and ULTRA-V (gray) endmembers for each pixel of the Samson data set.}
    \label{fig:samson_em_gray}
\end{figure}

To assess the estimated endmember variabilities, we analyzed the results for the Samson data set. We considered two approaches. The first approach consisted in averaging the projection of the estimated endmembers on the three eigenvectors associated to the three largest eigenvalues for each endmember. The results are shown in Fig.~\ref{fig:samson_em_pca}. These plots illustrate the endmember variances for each pixel, with red implying a large variance and blue a small variance. The second approach consisted in directly comparing the endmembers estimated with ULTRA-V and VCA. The results are shown in Fig.~\ref{fig:samson_em_gray}. These figures illustrate the ability of the proposed method to characterize the spectral variability while enforcing a spatial structure for the estimated endmembers.

\begin{figure}[htb]
    \centering
    \includegraphics[height=1.44in,width=0.48\textwidth]{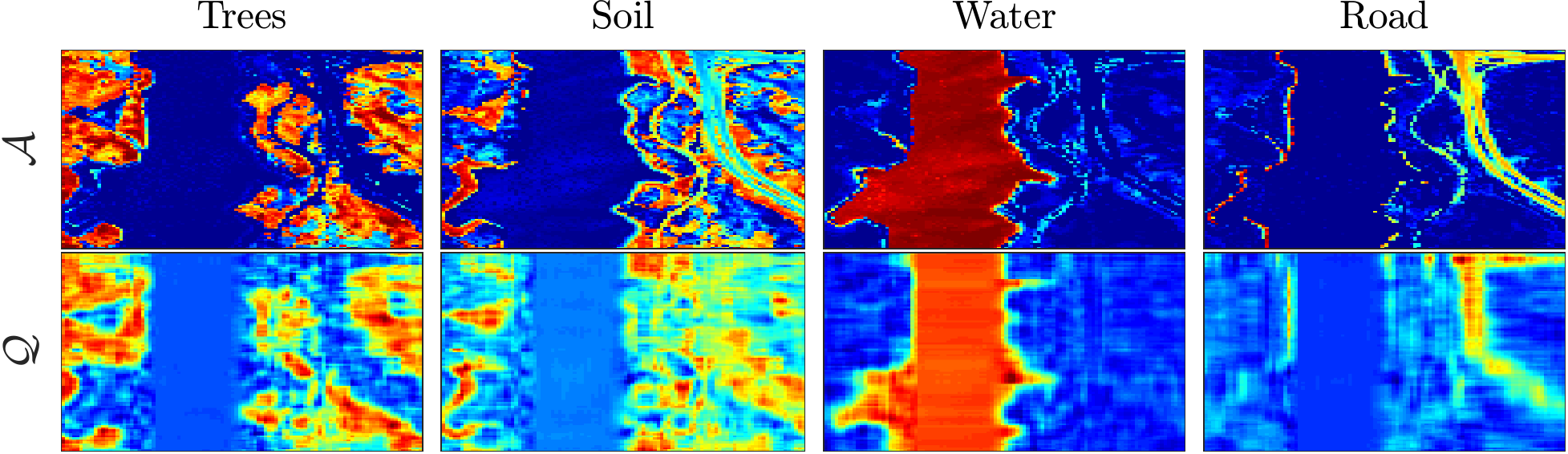}\\
    \caption{Comparison of tensors $\tensor{A}$ and $\tensor{Q}$ after ULTRA-V convergence for the Jasper Ridge data set.}
    \label{fig:jasper_AvsQ}
\end{figure}

\begin{figure}[htb]
    \centering
    \includegraphics[width=1.05\linewidth]{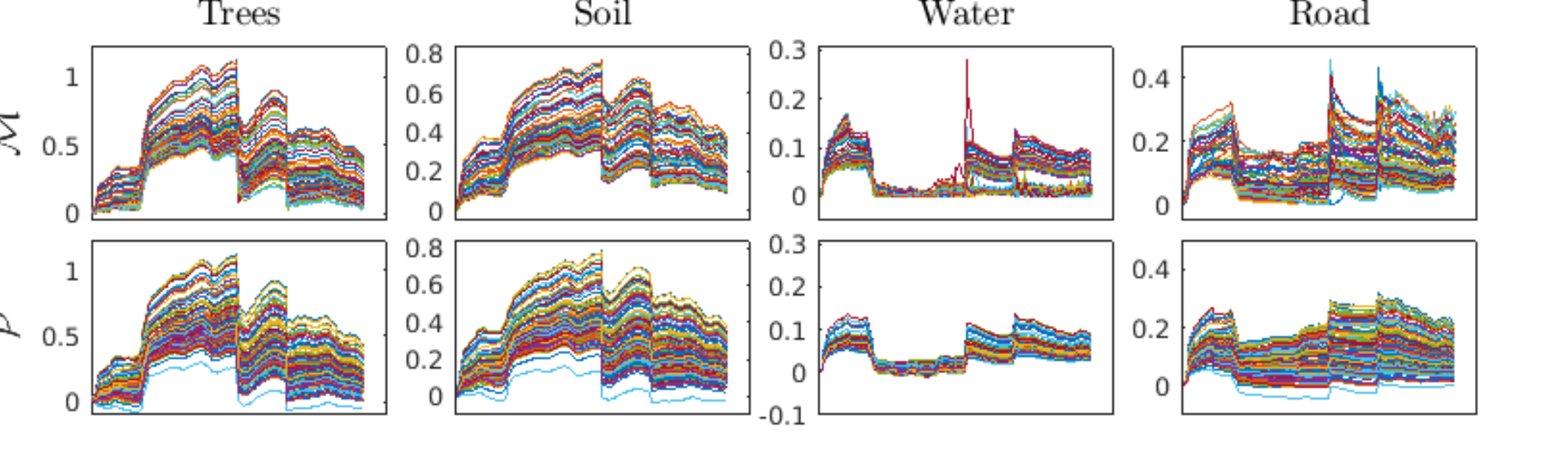}
    \caption{Comparison of tensors $\tensor{M}$ and $\tensor{P}$ after ULTRA-V convergence for the Jasper Ridge data set.}
    \label{fig:jasper_MvsP}
\end{figure}

To illustrate the role of the low-rank tensors $\tensor{P}$ and $\tensor{Q}$, we compare them to the abundances $\tensor{A}$ and endmembers $\tensor{M}$ in Figures~\ref{fig:jasper_AvsQ} and~\ref{fig:jasper_MvsP}, for the Jasper Ridge dataset. Figure~\ref{fig:jasper_AvsQ} shows the estimated abundances $\tensor{A}$ and their low-rank counterpart $\tensor{Q}$. One can verify that $\tensor{Q}$ (bottom row) has a very coarse spatial distribution when compared with $\tensor{A}$ (top row). This shows that imposing the low-rank structure through a regularization constraint gives the resulting abundances enough flexibility to model fine-scale spatial details while maintaining most of its spatial distribution. Figure~\ref{fig:jasper_MvsP} leads to similar conclusions for the endmembers. One can note that the low-rank tensor $\tensor{P}$ has a coarser structure when compared with the estimated endmembers $\tensor{M}$. This distinction can be seen very clearly for the Water and Road endmembers. 




\begin{table}[htb!]
\caption{Real Data.}
\centering
\renewcommand{\arraystretch}{1.2}
\resizebox{\linewidth}{!}{
\begin{tabular}{l|cc|cc|cc}
\toprule
Algorithm &\multicolumn{2}{c}{Houston Data}  & \multicolumn{2}{|c}{Samson} & \multicolumn{2}{|c}{Jasper Ridge}\\
\bottomrule\toprule
 & $\text{MSE}_{\tensor{R}}$ & Time & $\text{MSE}_{\tensor{R}}$ & Time & $\text{MSE}_{\tensor{R}}$ & Time \\ \midrule
FCLS	&	0.2283	&	1.90	&	0.0177	&	1.38	&	0.3567	&	1.59	\\
SCLS	&	0.0037	&	2.04	&	0.0041	&	1.29	&	0.0271	&	1.79	\\
PLMM	&	0.0190	&	454.90	&	0.0034	&	105.36	&	0.0257	&	72.86	\\
ELMM	&	\cblue{\bf{0.0010}}	&	474.45	&	7.82e-4	&	40.50	&	0.0058	&	100.49	\\
GLMM	&	\cred{\bf{1.0e-5}}	&	1326.85	&	\cred{\bf{0.2e-5}} &	50.62 & \cred{\bf{2.5e-5}} &	214.07	\\
ULTRA-V	&	0.0018	&	264.71	&	\cblue{\bf{6.4e-5}}	&	148.46	&	\cblue{\bf{15.0e-5}}	&	120.91	\\
\bottomrule\toprule
\end{tabular}}
\label{tab:results_realdata}
\end{table}

\section{Conclusions}\label{sec:conclusions}
%
In this paper, we proposed a new low-rank regularization strategy for introducing low-dimensional spatial-spectral structure into the abundance and endmember tensors for hyperspectral unmixing considering spectral variability. The resulting iterative algorithm, called ULTRA-V, imposes low-rank structure by means of regularizations that force most of the energy of the estimated abundances and endmembers to lay within a low-dimensional structure. The proposed approach does not confine the estimated abundances and endmembers to a strict low-rank structure, which would not adequately account for the complexity experienced in real-world scenarios. The proposed methodology includes also a strategy to estimate the rank of the regularization tensors $\tensor{P}$ and $\tensor{Q}$, leaving only two parameters to be adjusted within a relatively reduced search space. Simulation results using both synthetic and real data showed that the ULTRA-V can outperform state-of-the-art unmixing algorithms accounting for spectral variability.


\bibliographystyle{IEEEtran}
\bibliography{hyperspectral}

\vspace{-1cm}
\begin{IEEEbiography}[{\includegraphics[width=1in,height=1.25in,clip,keepaspectratio]{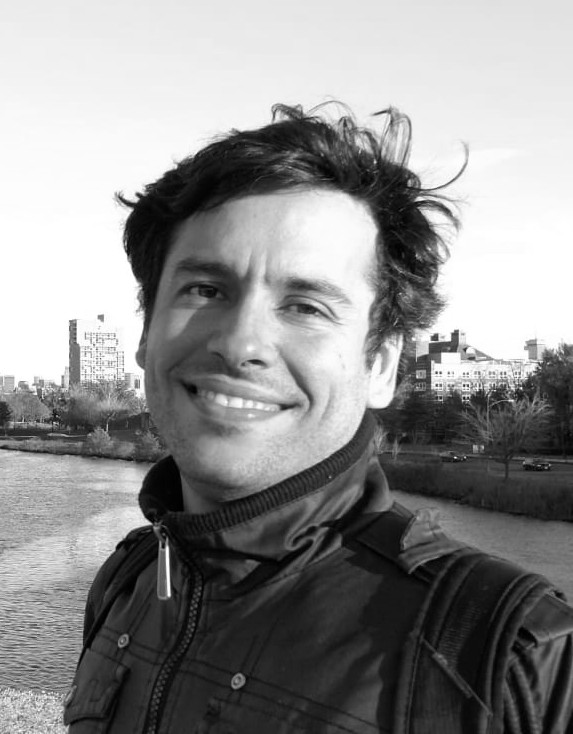}}]{Tales Imbiriba (S'14, M'17)}   
received his Doctorate degree from the Department of Electrical Engineering (DEE) of the Federal University of Santa Catarina (UFSC), Florian\'opolis, Brazil, in 2016. He served as a Postdoctoral Researcher at the DEE--UFSC and is currently a Postdoctoral Researcher at the ECE dept. of the Northeastern University, Boston, MA, USA. 
His research interests include audio and image processing, pattern recognition, kernel methods, adaptive filtering, and Bayesian Inference.
\end{IEEEbiography}

\begin{IEEEbiography}[{\includegraphics[width=1in,height=1.25in,clip,keepaspectratio]{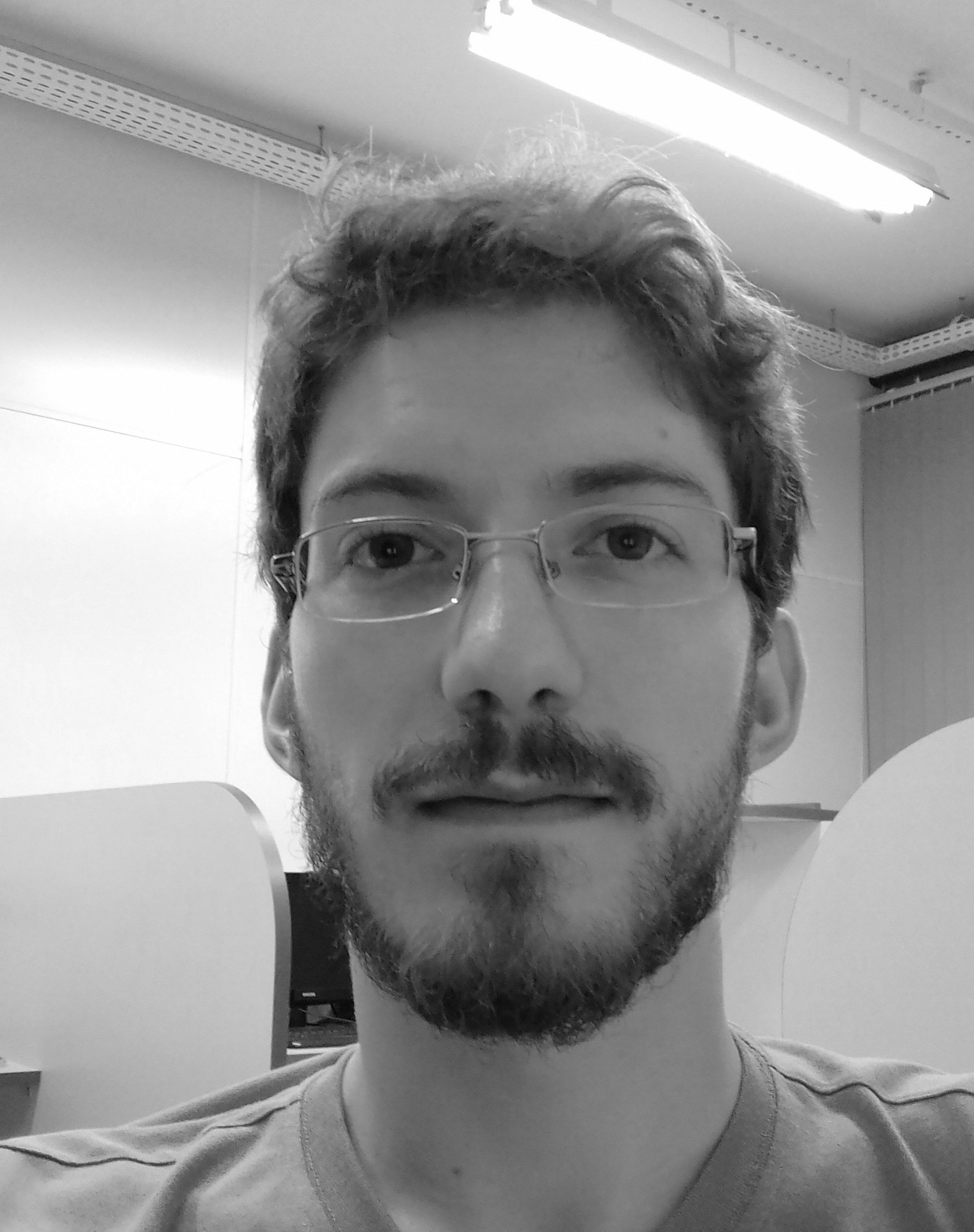}}]{Ricardo Augusto Borsoi (S'18)} 
received the MSc degree in electrical engineering from Federal University of Santa Catarina (UFSC), Florian\'opolis, Brazil, in 2016. He is currently working towards his doctoral degree at Universit\'e C\^ote d'Azur (OCA) and at UFSC. His research interests include image processing, tensor decomposition, and hyperspectral image analysis.
\end{IEEEbiography}

\vspace{-1cm}
\begin{IEEEbiography}[{\includegraphics[width=1in,height=1.25in,clip,keepaspectratio]{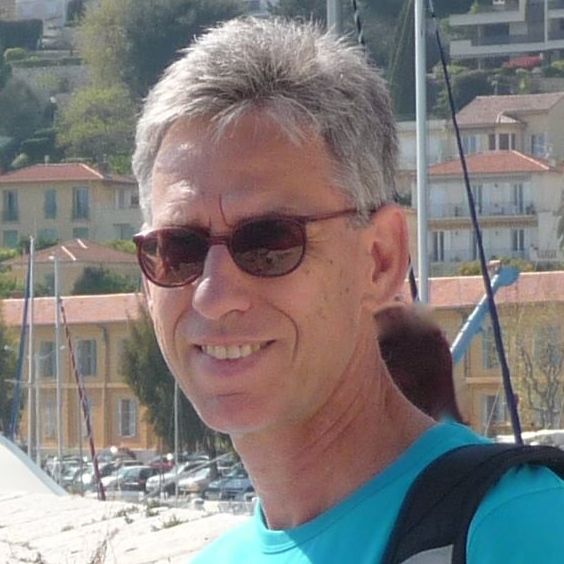}}]{Jos\'e Carlos M. Bermudez (S'78,M'85,SM'02)}
received the B.E.E. degree from the Federal University of Rio de Janeiro (UFRJ), Rio de Janeiro, Brazil, the M.Sc. degree in electrical engineering from COPPE/UFRJ, and the Ph.D. degree in electrical engineering from Concordia University, Montreal, Canada, in 1978, 1981, and 1985, respectively.
  He joined the Department of Electrical Engineering, Federal University of Santa Catarina (UFSC), Florianopolis, Brazil, in 1985. He is currently a Professor of Electrical Engineering at UFSC and a Professor at Catholic University of Pelotas (UCPel), Pelotas, Brazil. He has held the position of Visiting Researcher several times for periods of one month at the Institut National Polytechnique de Toulouse, France, and at Université Nice Sophia-Antipolis, France. He spent sabbatical years at the Department of Electrical Engineering and Computer Science, University of California, Irvine (UCI), USA, in 1994, and at the Institut National Polytechnique de Toulouse, France, in 2012. 
  His recent research interests are in statistical signal processing, including linear and nonlinear adaptive filtering, image processing, hyperspectral image processing and machine learning.
  Prof. Bermudez served as an Associate Editor of the IEEE TRANSACTIONS ON SIGNAL PROCESSING in the area of adaptive filtering from 1994 to 1996 and from 1999 to 2001. He also served as an Associate Editor of the EURASIP Journal of Advances on Signal Processing from 2006 to 2010, and as a Senior Area Editor of the IEEE TRANSACTIONS ON SIGNAL PROCESSING from 2015 to 2019.  He is the Chair of the Signal Processing Theory and Methods Technical Committee of the IEEE Signal Processing Society. Prof. Bermudez is a Senior Member of the IEEE.
\end{IEEEbiography}

\end{document}